\documentclass{article}

\usepackage{PRIMEarxiv}

\usepackage[utf8]{inputenc} 
\usepackage[T1]{fontenc}    
\usepackage{hyperref}       
\usepackage{url}            
\usepackage{booktabs}       
\usepackage{amsfonts}       
\usepackage{nicefrac}       
\usepackage{microtype}      
\usepackage{lipsum}
\usepackage{fancyhdr}       
\usepackage{graphicx} 
\graphicspath{{media/}}    
\usepackage{placeins}
\usepackage[numbers]{natbib}
\usepackage{amsmath}
\usepackage{breqn}
\usepackage{subcaption}
\usepackage{float}
\usepackage{rotating}
\usepackage[table]{xcolor}
\usepackage{xurl}   
\usepackage{hyperref}
\usepackage{comment}

\usepackage{tabularx}
\usepackage{array}
\usepackage{threeparttable}
\usepackage{makecell}
\renewcommand{\arraystretch}{1.15}

\title{
Integrating attention into explanation frameworks for language and vision transformers
}

\author{
  Marte Eggen, Jacob Lysnæs-Larsen, Inga Strümke \\
  Department of Computer Science \\
  Norwegian University of Science and Technology \\
  Trondheim, Norway \\
  \texttt{\{marte.eggen, jacob.lysnas-larsen, inga.strumke\}@ntnu.no} \\
}

\begin{document}

\FloatBarrier

\maketitle

\begin{abstract}
The attention mechanism lies at the core of the transformer architecture, providing an interpretable model-internal signal that has motivated a growing interest in attention-based model explanations. Although attention weights do not directly determine model outputs, they reflect patterns of token influence that can inform and complement established explainability techniques. This work studies the potential of utilising the information encoded in attention weights to provide meaningful model explanations by integrating them into explainable AI (XAI) frameworks that target fundamentally different aspects of model behaviour. To this end, we develop two novel explanation methods applicable to both natural language processing and computer vision tasks. The first integrates attention weights into the Shapley value decomposition by redefining the characteristic function in terms of pairwise token interactions via attention weights, thus adapting this widely used game-theoretic solution concept to provide attention-driven attributions for local explanations. The second incorporates attention weights into token-level directional derivatives defined through concept activation vectors to measure concept sensitivity for global explanations. Our empirical evaluations on standard benchmarks and in a comparison study with widely used explanation methods show that attention weights can be meaningfully incorporated into the studied XAI frameworks, highlighting their value in enriching transformer explainability.

\end{abstract}


\section{Introduction}

Transformer-based models have seen widespread adoption in a variety of fields, often surpassing other neural network architectures in handling sequential data. The advancements are particularly prominent in natural language processing (NLP) and computer vision, typically in models fine-tuned on supervised downstream tasks. Notable transformer-based architectures from these domains include the Bidirectional Encoder Representations from Transformers (BERT) and Vision Transformer (ViT), respectively \citep{devlin-etal-2019-bert, dosovitskiy2020image}. The former has improved performance in NLP applications such as text classification, whereas the latter has demonstrated the effectiveness of treating images as sequences of patches, being a compelling alternative to convolutional layers. 

As is common for neural networks, the decision process of transformer models is inherently non-interpretable to humans due to their non-linear transformations and high-dimensional embedding representations. Methods from the field of explainable AI (XAI) are developed to address this challenge. Central to the transformer architecture is the attention mechanism, which computes pairwise token relevance and captures long-range dependencies within sequential data. The attention weights provide a model-internal signal with a direct interpretation, thus constituting a natural foundation from which to develop model explanations. \citet{sundararajan2017axiomatic} advocate an axiomatic approach to explanation design by specifying formal properties that an attribution method should satisfy, motivating their method \emph{Integrated Gradients}, which provably adheres to said axioms. Following this view, fulfilling properties that are postulated a priori can be a favourable property of explanation methods. By extension, we argue that a desirable property of an explanation method is for it to make use of architectural mechanisms known to be central to the model. In particular, using attention weights as part of the explanation of a transformer model is justified when the goal is architectural faithfulness.

This work introduces two novel attention-based explanation methods that combine attention weights with diverse and well-established XAI frameworks, applicable to both NLP and computer vision tasks. While attention weights do not directly determine output contributions, they are known to encode explanatory information~\cite{abnar-zuidema-2020-quantifying, clark2019does, vig2019analyzing, wiegreffe-pinter-2019-attention}, and thus have unexplored potential when integrated into existing explainability approaches. The aim of this work is therefore to provide an empirical and methodological study of whether and how attention weights can meaningfully improve current explanation methods. 

The first proposed method uses attention weights to define the characteristic function for computing Shapley values \citep{shapley1953value} for local token attributions of individual predictions. The second method builds upon a concept-based explanation framework by adapting directional derivatives defined through concept activation vectors (CAVs) \citep{pmlr-v80-kim18d} to the token level in computing global concept sensitivity scores, capturing the presence of high-level concepts in the model's inner representations. 

In order to evaluate the proposed attribution method, we conduct a comparison study with two perturbation- and gradient-based approaches and pure attention weights for completeness. Further, in their survey on XAI methods for transformers, \citet{fantozzi2024explainability} stress the need for metrics in evaluating model explanations. In response to and agreement with this, we select three metrics for quantitatively evaluating and objectively comparing the produced explanations. Finally, for the concept-based method, we retain the experimental design used in prior work on concept-based explanation. 

This work makes the following contributions: 
\begin{itemize}
    \item Demonstrate the flexibility and utility of attention weights when incorporated into various well-established XAI frameworks for local and global explanations of transformer models for both NLP and computer vision tasks. 
    \item A comparison of feature importance methods adapted and applied to a transformer architecture for local explanations.
    \item Provide the ongoing debate regarding the explanatory power of the attention mechanism with use-cases and comparisons. 
\end{itemize}

The remainder of this study is structured as follows. Section \ref{sec:preliminaries} introduces the transformer architecture, the relevant theoretical background, and covers related work, with Sec.~\ref{sec:method} presenting in detail our two proposed attention-based explanation methods. Section \ref{sec:experimental_setup} outlines the experimental setup, including a description of the datasets, downstream tasks, and evaluation metrics, followed by the results presented in Sec.~\ref{sec:results}. Finally, Sec. \ref{sec:discussion_conclusion} summarises the main contributions and suggests possible directions for future research.

\section{Preliminaries}
\label{sec:preliminaries}
\subsection{The transformer architecture and attention mechanism 
\label{sec:transformer_architecture}}

The transformer model proposed by \citet{vaswani2017attention} comprises a stack of encoder and decoder layers, each applying a series of transformations to its input representations. The model processes an input sequence $x = (x_1, ..., x_N)$ of $N$ elements by first mapping each token $x_i$ for $i \in \{1, ..., N\}$ to a real-valued representation through an embedding function $E : V \rightarrow \mathbb{R}^d$, where $V$ denotes a pre-defined vocabulary and $d$ is the embedding dimension. The resulting embedded sequence $\mathbf{Z}^0_x = [\mathbf{e}_1^\top; ...; \mathbf{e}_N^\top] \in \mathbb{R}^{N \times d}$, where $\mathbf{e}_i = E(x_i)$, is augmented with positional encodings -- either learned or fixed -- to inform about the tokens' ordering. The subscript $x$ is omitted in the following to simplify notation. 

The attention mechanism is central to the transformer, allowing the model to weigh the relative importance of different sequence tokens. \citet{vaswani2017attention} introduce the notion of multi-head attention, where multiple attention heads $h \in \{1, ..., H\}$ operate in parallel, each with independent sets of learned queries, keys, and values.   

Given the latent representation $\mathbf{Z}^{\ell} \in \mathbb{R}^{N \times d}$ at layer $\ell \in \{1, ..., L\}$, the self-attention mechanism for head $h$ computes 
\begin{equation}
    \text{Attention}(\mathbf{Q}^{\ell h}, \mathbf{K}^{\ell h}, \mathbf{V}^{\ell h}) = \text{softmax}\left(\frac{\mathbf{Q}^{\ell h}{\mathbf{K}^{\ell h}}^\top}{\sqrt{d_k}}\right)\mathbf{V}^{\ell h},
\end{equation}
where the queries, keys, and values are obtained via learned linear projections,
\begin{equation}
    \mathbf{Q}^{\ell h} = \mathbf{Z}^{\ell}\mathbf{W}^{\ell h}_Q, \quad 
    \mathbf{K}^{\ell h} = \mathbf{Z}^{\ell}\mathbf{W}^{\ell h}_K, \quad \mathbf{V}^{\ell h} = \mathbf{Z}^{\ell}\mathbf{W}^{\ell h}_V,
\end{equation}
for $\mathbf{W}^{\ell h}_Q, \mathbf{W}^{\ell h}_K \in \mathbb{R}^{d \times d_k}$ and $\mathbf{W}^{\ell h}_V \in \mathbb{R}^{d \times d_v}$. The matrix $\mathbf{A}^{\ell h} = \text{softmax}\left(\frac{\mathbf{Q}^{\ell h}{\mathbf{K}^{\ell h}}^\top}{\sqrt{d_k}}\right) \in \mathbb{R}^{N \times N}$ provides the attention weights, determining the amount by which each token attends to itself and every other token in the sequence. 

The outputs of the attention heads in a layer $\ell$ are further concatenated and projected through a linear transformation $\mathbf{W}^{\ell}_O \in \mathbb{R}^{Hd_v \times d}$. The result is subsequently processed through a feed-forward network, with layer normalisation and residual connections applied after each attention and feed-forward sub-layer, yielding the updated latent representations $\mathbf{Z}^{\ell + 1}$. This general transformer architecture forms the foundation of a variety of models across different domains, including language and vision. 

BERT, a deep bidirectional transformer model for NLP designed to jointly condition on left and right contexts, is pre-trained on unlabeled text data, followed by task-specific fine-tuning \citep{devlin-etal-2019-bert}. ViT extends the transformer to computer vision tasks by treating images as sequences of non-overlapping patches. Specifically, an input image $\mathbf{X} \in \mathbb{R}^{I \times W \times C}$ is divided into a sequence of fixed-size patches $\mathbf{X}_p \in \mathbb{R}^{N \times (P^2 C)}$, where each patch has spatial resolution $(P, P)$ giving $N = IW / P^2$ patches \citep{dosovitskiy2020image}. The model is pre-trained on large-scale image data and subsequently fine-tuned with a classification head attached. Both architectures, BERT and ViT, comprise a stack of encoder layers and employ a special classification token $x_1 = x_{\text{CLS}}$ whose latent representation in $\mathbb{R}^d$ serves as an aggregated sequence summary used for downstream classification.

\subsection{Attention as explanation}
The attention mechanism has proven to be a highly effective component in neural networks, and a substantial portion of XAI research has investigated its potential for creating model explanations \citep{bibal-etal-2022-attention, fantozzi2024explainability}. 
Learned attention weights reflect token relations and may reveal broader patterns of the model's internal semantic or syntactic representations. \citet{vig2019analyzing} analyse the structure of attention in a transformer language model and find that attention aligns most strongly with syntactic dependency relations in the middle model layers. Early layers tend to focus on positional information, while deeper layers capture more specific linguistic constructs and longer-distance relationships. Similarly, \citet{clark2019does} study attention in BERT and observe that some attention heads adhere to interpretable patterns, such as attending to fixed positional offsets or particular syntactic relations. No single head captures several relations, indicating that linguistic knowledge is distributed across multiple attention heads. Notably, the literature also presents studies challenging the 
explanatory power of attention weights \citep{bibal-etal-2022-attention, jain-wallace-2019-attention}, which has given rise to a debate regarding their direct utility as model explanations. The interested reader is referred to \citet{bibal-etal-2022-attention} for an introduction to this debate. 

While attention-based model explanations often involve processing the attention weights, the potential for systematically integrating them into well-established XAI frameworks remains underexplored. An exception is the Gradient Self-Attention Maps (Grad-SAM) method proposed by \citet{barkan2021grad}, which augments gradient-based model explanations by incorporating attention weights. They compute local feature attribution scores based on the Hadamard product of ReLU-activated gradients, with respect to the attention weights, and the weights themselves.

\subsection{Shapley value based explanations}
\label{sec:shapley_values}

The Shapley decomposition was originally introduced as a solution concept in cooperative game theory to fairly distribute a game's total payoff among the individual players participating in the game \citep{shapley1953value}. Formally, given a set of players $\mathcal{N} = \{1, ..., N\}$ and a characteristic function $v : 2^{\mathcal{N}} \rightarrow \mathbb{R}$ that assigns a value to each coalition of players $\mathcal{S} \subseteq \mathcal{N}$, the Shapley value for a player $i \in \mathcal{N}$ is defined as

\begin{equation}
    \phi_i(v) = \sum_{\mathcal{S} \subseteq \mathcal{N} \setminus \{i\}} \frac{|\mathcal{S}|! (N - |\mathcal{S}| - 1)!}{N!} \left( v(\mathcal{S} \cup \{i\}) - v(\mathcal{S}) \right),
    \label{eq:shapley_values}
\end{equation}
where $|\mathcal{S}|$ denotes the size of coalition $\mathcal{S}$ and $v(\emptyset) = 0$.

Equation~\eqref{eq:shapley_values} computes a weighted average of the player's marginal contribution $v(\mathcal{S} \cup \{i\}) - v(\mathcal{S})$ across all possible coalitions $\mathcal{S}$. The Shapley value is provably the only decomposition satisfying the favourable properties of efficiency, symmetry, null player, and linearity \citep{young1985monotonic}. 

The Shapley decomposition can be applied to neural networks by treating the model's output predictions across all combinations of input features (analogous to players) as coalition values. For a large number of features, it is computationally infeasible to calculate exact Shapley values, as the number of terms in Eq.~\eqref{eq:shapley_values} grows exponentially with $N$. A common approximation technique is to sample a subset of coalitions, either randomly or through specific sampling strategies. In KernelSHAP \citep{lundberg2017unified, olsen2024improving}, coalitions are sampled from a distribution
\begin{equation}
    k(N, |\mathcal{S}|) = \frac{N - 1}{{N\choose |\mathcal{S}|} |\mathcal{S}| (N - |\mathcal{S}|)},
    \label{eq:kernelshap}
\end{equation}
for coalition sizes $|\mathcal{S}| \in \{0, ..., N\}$. Notably, $k(N, 0) = k(N, N) = \infty$, implying that the empty and grand coalitions are always included \citep{olsen2024improving}. 

SHapley Additive exPlanations (SHAP) \citep{lundberg2017unified} is a well-established XAI framework for local perturbation-based model explanations based on the Shapley decomposition. In this approach, a deep neural network is approximated with a simpler explanation model, defined as ``any interpretable approximation of the original model'' \citep{lundberg2017unified}. The generated SHAP values represent the change in the expected model prediction when conditioning on a particular feature, explaining the shift from a base value -- being the expected output without the feature information -- to the model prediction for the observed input feature. Moreover, SHAP values fulfil the aforementioned desirable properties of the Shapley value \citep{lundberg2017unified}.

Several studies relate both Shapley and SHAP values to attention weights and transformer models. \citet{ethayarajh-jurafsky-2021-attention} show that attention flows, a post-processed version of attention weights derived from applying the max-flow algorithm to the attention graph, are themselves Shapley values. \citet{kokalj-etal-2021-bert} propose TransSHAP, a method adapting SHAP to transformer models. \citet{sun2023efficient} further compute token attributions by both decoupling the self-attention from the transformer's information flow and freezing unrelated values to reformulate the transformer layer output to a linear function. Treating the attention paid to each token as an individual contributor -- akin to a player in a collaborative game -- they obtain token attributions based on how the presence or absence of attention to each token impacts the output.

\subsection{Concept-based explanations} 
\label{sec:concept-based_explanations}

Concept-based XAI aims to explain neural networks by identifying whether human-understandable, abstract concepts in the input are represented as higher-level representations across the model's latent layers. A common approach involves training an, often linear, probe on these representations to assess its ability to distinguish between pre-defined concepts \citep{pmlr-v80-kim18d}.

Assume a model input $\mathbf{x} \in \mathbb{R}^n$ processed at an intermediate layer $\ell$ with $m$ neurons that yield activations described by a function $f^{\ell} : \mathbb{R}^n \rightarrow \mathbb{R}^m$. A linear classifier probe is trained to discriminate between sets of activations $\{f^{\ell}(\mathbf{x}) : \mathbf{x} \in P_C\}$ and $\{f^{\ell}(\mathbf{x}) : \mathbf{x} \in N_{C}\}$, where $P_C$ is a set of positive examples related to a concept $C$ and $N_{C}$ a negative set of, e.g., random examples. This binary classification task assesses the degree to which these activations encode the concept $C$ in a linearly separable manner. The associated CAV is defined as a unit vector $\mathbf{v}_C^{\ell} \in \mathbb{R}^m$, the direction in the activation space orthogonal to the decision boundary of the trained classifier probe \citep{pmlr-v80-kim18d}. 

The concept sensitivity, being the change in the model output along the direction of the CAV for a class $k$ and concept $C$ at layer $\ell$, is measured as the scalar directional derivative,
\begin{equation}
    S_{C, k}^{\ell}(\mathbf{x}) = \nabla g_{k}^{\ell} \left( f^{\ell}(\mathbf{x})\right) \cdot \mathbf{v}_C^{\ell},
    \label{eq:directional_derivative}
\end{equation}
where $g_{k}^{\ell} : \mathbb{R}^m \rightarrow \mathbb{R}$ denotes the remaining layers of the network and $\nabla$ denotes the gradient \citep{pmlr-v80-kim18d}. 

\citet{pmlr-v80-kim18d} introduce Testing with CAVs (TCAV), a method that uses directional derivatives to quantify concept sensitivity across a set of input samples $X_k$ from a target class $k$, 

\begin{equation}
    \text{TCAV}^{\ell}_{C, k} = \frac{|\{\mathbf{x} \in X_k : S_{C, k}^{\ell}(\mathbf{x}) > 0 \}|}{|X_k|} \in [0,1],
    \label{eq:tcav}
\end{equation}

representing the fraction of inputs in $X_k$ for which the concept $C$ positively influences the activations at layer $\ell$. Notably, the TCAV score only depends on the sign of the directional derivative $S_{C, k}^{\ell}(\mathbf{x})$, rather than its magnitude.

Relative CAVs are constructed by directly comparing multiple semantically meaningful concepts. Rather than contrasting a concept $C$ with a generic negative set $N_{C}$, the relative CAV, denoted $\mathbf{v}_{C,D}^{\ell} \in \mathbb{R}^m$, represents a direction in the activation space that separates the concept $C$ from the (possibly composite) concept $D$ \citep{pmlr-v80-kim18d}. This relative CAV may replace the standard $\mathbf{v}_C^{\ell}$ in computing directional derivatives and TCAV scores in Eqs.~\eqref{eq:directional_derivative} and \eqref{eq:tcav}, respectively. 

Developments in concept-based XAI for transformers include the ConceptTransformer, presented by \citet{rigotti2021attention}, a module designed to replace the conventional classification head to provide both faithful and plausible model explanations. This module leverages cross-attention between input representations and embeddings encoding domain-specific concepts. \citet{sinha2025ascentvit} also introduce a concept-learning framework integrated as a classification head for a ViT backbone. The approach uses scale and position-aware vector representations, which are further aligned with concept annotations through attention computation. Another line of work explores a post-hoc, concept-based explanation method for NLP classification tasks, as proposed by \citet{jourdan2023cockatiel}. They employ non-negative matrix factorisation to discover concepts influencing model predictions, followed by a sensitivity analysis to estimate each concept's contribution to the output.

\section{Explanation methods}
\label{sec:method}

\subsection{Attention-based attributions through the Shapley decomposition}
\label{sec:shaptention}

The first method we propose follows the notion of the Shapley decomposition, treating all input features as players in a collaborative game with transferable utility. Such a game is completely defined by a finite set of players and the characteristic function assigning a value to each coalition of players. In the context of transformer models, faithful token attributions can be derived by defining the input tokens as players and using the model's prediction to form the characteristic function. However, this approach requires multiple forward passes to evaluate the model for all possible coalitions. To reduce the cost, an alternative characteristic function can be defined to be more computationally tractable. Upon changing the characteristic function, the game must be reinterpreted accordingly, giving a proxy game approximating the original.

Attention weights reflect relational dependencies between tokens, thereby forming a natural basis from which to construct a characteristic function when treating tokens as players in a cooperative game. Notably, attention weights themselves provably cannot be Shapley values~\citep{ethayarajh-jurafsky-2021-attention}; however, they can still form part of a collaborative game by contributing systematically to the characteristic function. 

To this end, we construct characteristic functions based on attention weights in simplified games. For a simplified game to yield meaningful token attributions, it must approximate the original model behaviour with sufficient fidelity. I.e., the resulting Shapley values correspond to output contributions which are faithful to the proxy characteristic function and approximate the true behaviour only insofar as the proxy faithfully captures the relevant dynamics of the original model. A first-order Taylor expansion can be formulated to establish the relationship between internal attention dynamics and the model output. Assume a scalar-valued component function $g_k^{\ell h} : \mathbb{R}^{N \times N} \rightarrow \mathbb{R}$ of the full model, mapping from the intermediate attention matrix $\mathbf{A}^{\ell h}$ at layer $\ell$ and head $h$ to the output logit for class $k$. The first-order Taylor approximation around a zero baseline is 
\begin{equation}
    g_k^{\ell h}(\mathbf{A}^{\ell h}) \approx \sum_{i=1}^N \sum_{j = 1}^N A_{ij}^{\ell h} \cdot \frac{\partial g_k^{\ell h}}{\partial A_{ij}^{\ell h}},
    \label{eq:taylor_approx}
\end{equation}
expressing the contribution of the attention weights $\mathbf{A}^{\ell h}$ to the model output when isolating their influence, under the assumption of local linearity around the current value. 

Following \citet{barkan2021grad}, we define the aggregate contribution across all layers and heads as the average of their Hadamard products,
\begin{equation}
    \mathbf{M}_k = \frac{1}{LH} \sum_{\ell = 1}^{L} \sum_{h = 1}^H \mathbf{A}^{\ell h} \circ \text{ReLU}(\mathbf{G}_k^{\ell h}) \in \mathbb{R}^{N \times N},
    \label{eq:sum_hadamard_prod}
\end{equation}
where $\mathbf{G}_k^{\ell h} \in \mathbb{R}^{N \times N}$ denotes the gradients of the output logit for class $k$ with respect to $\mathbf{A}^{\ell h}$. Applying the ReLU function to the gradients as in Eq.~\eqref{eq:sum_hadamard_prod} modifies the classical first-order Taylor approximation by thresholding negative derivatives to zero, focusing exclusively on contributions that actively promote the output logit. To obtain per-token attributions, entries in $\mathbf{M}_k$ must be aggregated, which we suggest to achieve through the Shapley decomposition. In the following, three alternative formulations of the characteristic function are proposed for computing Shapley values, each defined solely in terms of values in $\mathbf{M}_k$.

\paragraph{Attention interactions with the classification token}
The first proposed characteristic function focuses exclusively on the attention interactions between the input tokens and the classification token $x_{\text{CLS}}$, whose latent representation drives the model prediction. Given an input sequence $x$ including special tokens such as $x_{\text{CLS}}$, let $x_P$ denote the subsequence of original (non-special) tokens, with corresponding index set $\mathcal{X}_P$. The value of a coalition $\mathcal{S} \subseteq \mathcal{X}_P$ is defined through the sum of attention weights from the classification token to tokens in $x_P$,

\begin{equation}
    v(\mathcal{S}) = \sum\limits_{i \in \mathcal{S}} M_{k,\text{CLS}i},
    \label{eq:char_func_grad_att_cls}
\end{equation}
where the scalar entries $M \in \mathbb{R}$ originates from $\mathbf{M}_k$ in Eq.~\eqref{eq:sum_hadamard_prod} and $v(\emptyset) = 0$.

\paragraph{Mutual attention interactions}
Rather than relying on token importance as perceived by the classification token across all layers and heads, the second proposed function is based on the attention weights capturing mutual interactions among original input tokens. The value of a coalition $\mathcal{S} \subseteq \mathcal{X}_P$ is then defined using pairwise contribution scores 
\begin{equation}
    v(\mathcal{S}) =     
    \begin{cases}
      \sum\limits_{\substack{i, j \in \mathcal{S} \\ i \neq j}} \left( M_{k, ij} + M_{k, ji} \right), & \text{if } |\mathcal{S}| > 1 \\
      M_{k, ii}, & \text{if } \mathcal{S} = \{i\} \\
      0, & \text{otherwise} \\
    \end{cases}.
    \label{eq:char_func_grad_att_mutual}
\end{equation}

Intuitively, the Shapley values obtained from this characteristic function emphasise the tokens' contributions in the context of their relations with all other tokens through the attention mechanism. 

An alternative strategy would involve additionally including the relevant diagonal elements of $\mathbf{M}_k$ in coalitions for which $|\mathcal{S}| > 1$. However, these terms primarily reflect a token's intrinsic importance to itself, rather than its contribution to a coalition. By focusing solely on pairwise or bidirectional interactions as in Eq.~\eqref{eq:char_func_grad_att_mutual}, the intention is to better capture the relational influence between tokens, being more consistent with the collaborative game setting with transferable utility, for which the Shapley decomposition was developed. 

\paragraph{Maximum mutual attention interactions}
The third proposed characteristic function follows the previous, while introducing non-linearity through the max function,

\begin{equation}
    v(\mathcal{S}) =     
    \begin{cases}
      \sum\limits_{\substack{i, j \in \mathcal{S} \\ i \neq j}} \text{max}\left( M_{k, ij}, \, M_{k, ji} \right), & \text{if } |\mathcal{S}| > 1 \\
      M_{k, ii}, & \text{if } \mathcal{S} = \{i\} \\
      0, & \text{otherwise} \\
    \end{cases},
    \label{eq:char_func_grad_att_mutual_max}
\end{equation}
where $\mathcal{S} \subseteq \mathcal{X}_P$.

We present Eqs.~\eqref{eq:char_func_grad_att_cls}, \eqref{eq:char_func_grad_att_mutual} and \eqref{eq:char_func_grad_att_mutual_max} as characteristic functions for computing Shapley values according to Eq.~\eqref{eq:shapley_values} for each token in $x_P$. The structure of the proposed characteristic functions merits further comments:

The integration of attention weights into the Shapley framework as suggested in Eq.~\eqref{eq:char_func_grad_att_cls}, where the characteristic function is defined as a pure sum, causes a constant contribution from each player to every coalition. This would yield a systematic cancellation in the last term of Eq.~\eqref{eq:shapley_values}, rendering the resulting Shapley values merely a scaled version of the original values $M_{k, \text{CLS}i}$ for all tokens $x_i \in x_P$. Similarly, the Shapley formula in Eq.~\eqref{eq:shapley_values} with characteristic function as defined by Eq.~\eqref{eq:char_func_grad_att_mutual} can be reduced to an expression with non-exponential complexity due to its linearity. This derivation is provided in App.~\ref{app:derivation_shapley_mutual_attn}. A similar simplification does not apply to Eq.~\eqref{eq:char_func_grad_att_mutual_max} due to the non-linear max function. However, approximation strategies discussed in Sec.~\ref{sec:shapley_values} can be applied to sample coalitions in order to estimate the exact values within feasible computational time for large player sets. The selected coalitions are subsequently used to compute attributions according to Eq.~\eqref{eq:shapley_values}.

The Shapley value approach using characteristic functions defined from $\mathbf{M}_k$ offers a theoretically grounded alternative to a na\"ive aggregation of these values into token attributions, retaining the desirable properties of Shapley values. The resulting per-token Shapley values from all formulations provided in Eqs.~\eqref{eq:char_func_grad_att_cls}, \eqref{eq:char_func_grad_att_mutual}, and \eqref{eq:char_func_grad_att_mutual_max} are considered token attributions following Eq.~\eqref{eq:taylor_approx}. However, all versions correspond to different games depending on the definition of the characteristic function, and implicitly reflect different assumptions and approximations about the model's internal behaviour. Moreover, using $\mathbf{M}_k$ as the basis for token attributions assumes that the attention mechanism is the main mediator of output-relevant information flow and that the product of attention weights and their positive gradients serves as a reasonable estimate for the tokens' influence on the model prediction. 

We assess and compare the performance of the methods outlined in this section against established local explanation methods for feature attribution. The comparison includes the widely used perturbation-based method SHAP \citep{lundberg2017unified}, as well as the gradient-based approach Grad-SAM \citep{barkan2021grad}. Attention weights alone do not provide faithful feature attributions, as they may not directly correlate with the model output. Nonetheless, we include versions of the methods presented in this section using only raw attention weights without gradient information to illustrate their behaviour in isolation, see App.~\ref{app:char_func_shapley_att}. A summary of all methods included in the comparison is listed in Tab.~\ref{tab:methods_nlp}. 

\begin{table}
   \centering
   \renewcommand{\arraystretch}{1.5}
   \caption{
   The complete list of explanation methods evaluated on the NLP classification tasks defined in Sec.~\ref{sec:experimental_setup}.
   \label{tab:methods_nlp}}
    \resizebox{\textwidth}{!}{
   \begin{tabular}{p{2cm}p{14cm}p{2cm}}
       \toprule
       \textbf{Method} & \textbf{Description} & \textbf{Reference}\\
       \midrule
       Random & Input token importances sampled from a uniform distribution. & - \\
       Shapley-Att-CLS &  Input token importances computed as Shapley values over attention weights from the classification token to original input tokens. & Eq.~\eqref{eq:char_func_att_cls} \\
       Shapley-Att-Mutual & Input token importances computed as Shapley values over attention weights that reflect mutual attention interactions among tokens in the player coalition. & Eq.~\eqref{eq:char_func_att_mutual} \\
       Shapley-Att-Max-Mutual & Input token importances computed as Shapley values over attention weights that reflect the strongest mutual interaction between each token pair in the player coalition. & Eq.~\eqref{eq:char_func_att_mutual_max} \\
       Rand Shapley-Att-Max-Mutual & An approximation of Shapley-Att-Max-Mutual with random sampling of player coalitions. & Sec.~\ref{sec:shapley_values} \\
       Kernel Shapley-Att-Max-Mutual & An approximation of Shapley-Att-Max-Mutual with player coalitions sampled according to Eq.~\eqref{eq:kernelshap}. & Sec.~\ref{sec:shapley_values} \\

       Grad-SAM & Input token importances computed as the sum of all attention weights, multiplied element-wise by the ReLU-activated gradient of the output with respect to the corresponding attention weights, across all layers and heads. & \citet{barkan2021grad} \\
       Shapley-Grad-Att-CLS & Input token importances quantifying each token's contribution to the model output through its interactions with the classification token in the attention mechanism, computed as Shapley values over attention-weighted gradients. & Eq.~\eqref{eq:char_func_grad_att_cls}\\
       Shapley-Grad-Att-Mutual & Input token importances quantifying the contribution to the model output via token interactions in the attention mechanism, computed as Shapley values over attention-weighted gradients that reflect mutual interactions among tokens in the player coalition. & Eq.~\eqref{eq:char_func_grad_att_mutual}\\
       Shapley-Grad-Att-Max-Mutual & Input token importances quantifying the contribution to the model output via token interactions in the attention mechanism, computed as Shapley values over attention-weighted gradients that reflect the strongest mutual interaction between each token pair in the player coalition. & Eq.~\eqref{eq:char_func_grad_att_mutual_max}\\
       Rand Shapley-Grad-Att-Max-Mutual & An approximation of Shapley-Grad-Att-Max-Mutual with random sampling of player coalitions. & Sec.~\ref{sec:shapley_values} \\
       Kernel Shapley-Grad-Att-Max-Mutual & An approximation of Shapley-Grad-Att-Max-Mutual with player coalitions sampled according to Eq.~\eqref{eq:kernelshap}. & Sec.~\ref{sec:shapley_values} \\
       
       Shapley-Input & Input token importances for the model prediction by the Shapley decomposition. & Eq.~\eqref{eq:char_func_shapley_input}\\
       SHAP & Perturbation-based input token importances quantifying the contributions to the final model prediction's deviation from the expected prediction across a background dataset. & \citet{lundberg2017unified} \\

       \bottomrule 
   \end{tabular}
   }
\end{table}

\subsection{Concept-based directional derivatives through attention}
\label{sec:attn_directional_derivative}
We further propose an adaptation of the TCAV score introduced by \citet{pmlr-v80-kim18d} that incorporates attention weights. At a high level, the transformer operates within a consistently dimensioned space across all layers, implying that each layer's output token representation resides in $\mathbb{R}^d$. \citet{elhage2021mathematical} refer to a ``residual stream'', conceptualised as a communication channel where individual transformer modules interact within the shared high-dimensional vector space. The CAV is to be understood as a direction in this latent space for a specific concept. Since the latent token representations also exist in this space, this allows for an equivalent directional derivative as in Eq.~\eqref{eq:directional_derivative} at the token level. Aligning token-level representations with CAVs in the shared vector space bridges the gap between global concept information and local feature relevance. 

Furthermore, the variance across tokens of concept sensitivity on the one hand, and of the attention weights from the classification token to the image tokens on the other, shows mirroring trends. In early layers, as shown in Figs.~\ref{fig:variance_all_classes} and~\ref{fig:variance_dalmatian}, the classification token attends approximately uniformly to all other tokens. The opposite is observed in deeper layers, where the attention distribution becomes increasingly concentrated, with a subset of tokens receiving substantially more weight. This suggests a progressive refinement of the model's focus toward tokens more relevant for classification. In contrast, concept sensitivity as measured using directional derivatives shows the opposite trend. The initial high variance decreases with layer depth, indicating that concept-related information is no longer localised. We hypothesise that this behaviour is due to communication via self-attention, in which concept information is shared between tokens. As a result, even tokens without a direct correspondence to the concept, encode concept-relevant signals in deeper layers. We exploit this complementary behaviour by weighting token-wise concept sensitivity with the classification token's attention weights. This informs the resulting explanation about tokens the model attended to during the forward computation, thereby reflecting the model’s internal information flow. When concept sensitivity is uniformly distributed, attention weights provide the necessary selectivity to identify which concept-bearing tokens have the most influence on the output. Thus, the added information, namely the increasing selectivity of attention weights, counterbalances the increasing uniformity of concept sensitivity in deeper layers, rendering the total score more informative for an architecture using the attention mechanism. This motivates a method combining the two, as described in the following.

A logistic regression classifier is trained on the model's latent representations of all tokens $\mathbf{z}_{i}^{\ell} \in \mathbb{R}^d \; \forall i \in \{1, ..., N\}$ yielding a classification hyperplane in the latent space. Its normal vector is normalised to obtain a unit CAV $\mathbf{v}_C^{\ell} \in \mathbb{R}^d$ associated with concept $C$ at model layer $\ell$, where $d$ is the transformer embedding dimension. We define a token-level, attention-weighted directional derivative as 

\begin{equation}
    \text{S}_{C, k, i}^{\ell, \text{attn}}(x) = \left(\frac{1}{H} \sum_{h=1}^H A_{\text{CLS}, i}^{\ell h} \right) [\nabla g_k^{\ell} (f^{\ell}(x))]_i  \cdot \mathbf{v}_C^{\ell},
    \label{eq:token_directional_derivative}
\end{equation}

where $f^{\ell}$ denotes the mapping from the input sequence $x$ to the latent states at layer $\ell$, and $g_k^{\ell}$ maps those latent representations to the scalar logit corresponding to class $k$. The notation $[\cdot]_i$ extracts the gradient corresponding to the $i$-th token, and the scalar $A_{\text{CLS}, i}^{\ell h}$ defines the attention weight of the classification token attending to the $i$-th token. Observe that the attention component in Eq.~\eqref{eq:token_directional_derivative} can be interpreted as a weight for each token's concept sensitivity. 

These directional derivatives preserve the spatial resolution of the input by producing a separate value for each token. This directly enables finer-grained maps localising concept sensitivity, rather than collapsing information into a single globally pooled score. However, in order to compute a scalar score analogous to the TCAV score in Eq.~\eqref{eq:tcav}, the per-token directional derivatives must be aggregated accordingly,

\begin{equation}
    S_{C, k}^{\ell, \text{attn}}(x) = \frac{1}{N} \sum_{i=1}^N S_{C, k, i}^{\ell, \text{attn}}(x).
\end{equation}

The adapted per-token TCAV score is then computed equivalently to Eq.~\eqref{eq:tcav} as

\begin{equation}
    \text{T-TCAV}_{C, k}^{\ell} = \frac{|x \in X_k : S_{C, k}^{\ell, \text{attn}}(x) > 0 \}|}{|X_k|} \in [0, 1],
\end{equation}

where $X_k$ is a set of input samples from class $k$. The T-TCAV score can be computed per model layer, as attention patterns are expected to vary throughout the network. This variation occurs because the effective influence between tokens accumulates across layers, sometimes referred to as attention flow \citep{abnar-zuidema-2020-quantifying}.

\section{Experiments and datasets}
\label{sec:experimental_setup}

\subsection{Datasets and downstream tasks}
\label{sec:datasets}
The methods from Sec.~\ref{sec:shaptention}, summarised in Tab.~\ref{tab:methods_nlp}, are evaluated on BERT models fine-tuned for both binary and multiclass text classification tasks with default settings\footnote{\url{https://huggingface.co/textattack/bert-base-uncased-SST-2}, \url{https://huggingface.co/textattack/bert-base-uncased-imdb}, \url{https://huggingface.co/textattack/bert-base-uncased-ag-news}}. The evaluation is conducted on the three datasets \textit{The Stanford Sentiment Treebank (SST-2)} \citep{socher-etal-2013-recursive}, \textit{IMDb} \citep{maas-etal-2011-learning}, and \textit{Ag News} \citep{NIPS2015_250cf8b5}, of which the two former cover a sentiment analysis task for binary classification, while the latter consists of news articles categorised into pre-defined topics for multiclass classification. 

Computing exact Shapley values according to Eq.~\eqref{eq:shapley_values} is computationally infeasible for large sets of players. To allow for exact computation even for methods with exponential computational cost, we extract a subset of the SST-2 test samples containing 15 tokens or fewer, referred to as \textit{SST-2 short}. In contrast, the IMDb and Ag News datasets typically consist of considerably longer text paragraphs, requiring approximations of such methods. We approximate by sampling 100 coalitions per data point as indicated in Tab.~\ref{tab:methods_nlp}.

The method presented in Sec.~\ref{sec:attn_directional_derivative} is evaluated using a ViT model fine-tuned for multi-class image classification with default settings\footnote{\url{https://huggingface.co/google/vit-base-patch16-224}}. The data used in the experiments are drawn from ImageNet \citep{5206848} and the Broden dataset \citep{8099837}, with the selected target classes and corresponding concepts primarily retrieved from the experiments by \citet{pmlr-v80-kim18d}.

Following \citet{pmlr-v80-kim18d}, we compute relative CAVs by training a logistic regression classifier to distinguish each concept from others. For each target concept, we use 120 concept-related images for the positive set, while the negative set is constructed by randomly sampling 120 images from the union of the negative concept sets. The T-TCAV scores are computed using 200 random samples from the ImageNet class datasets corresponding to the selected target classes. The directional derivatives, obtained by combining the relative CAVs with the gradients, are scaled to unity.

\subsection{Explanation metrics and evaluation}
\label{sec:metrics}

The explanation methods presented in Tab.~\ref{tab:methods_nlp} are quantitatively evaluated using the three metrics F1 score, comprehensiveness, and sufficiency, briefly described in the following.

The \textbf{$\mathbf{F_1}$ score} is computed using a reduced input $r_{:b\%}$, created by preserving the top-$b\%$ important tokens obtained by sorting the importance scores in descending order, while masking the remaining tokens \citep{barkan2021grad}. The model prediction on the reduced input, $f(r_{:b\%})$, is compared to the original prediction $f(x)$. The F$_1$ score is then calculated as
\begin{equation}
    \text{F}_1 \, \text{score} = \frac{2\text{TP}}{2\text{TP} + \text{FP} + \text{FN}}.
    \label{eq:f1}
\end{equation}
Specifically, we report the weighted F$_1$ score, as it is well-suited for multi-class classification problems. As an XAI metric, a higher F$_1$ score indicates more faithful feature attributions. 

\textbf{Comprehensiveness}, as defined by \citet{ferrando-etal-2022-measuring}, quantifies the change in the model's output prediction after removing important tokens,
\begin{equation}
    \text{Comprehensiveness} = \frac{1}{|B| + 1} \sum_{b \in B} \left( f_k(x) - f_k(x \backslash r_{:b\%}) \right),
    \label{eq:comp}
\end{equation}
where $r_{:b\%}$ defines the top-$b\%$ most important tokens and the function $f_k$ denotes the model's output for class $k$ predicted from the full, unmasked input. An attribution method performing well in identifying important tokens yields a large prediction difference, thereby achieving a higher comprehensiveness score.

\textbf{Sufficiency}, as defined by \citet{ferrando-etal-2022-measuring}, measures whether important tokens alone can retain the original model prediction, 
\begin{equation}
    \text{Sufficiency} = \frac{1}{|B| + 1} \sum_{b \in B} \left( f_k(x) - f_k(r_{:b\%}) \right).
    \label{eq:suff}
\end{equation}
A faithful explanation is characterised by low sufficiency values, as the most relevant tokens would better restore the original prediction, leading to a smaller prediction difference.

A fixed value of $b = 20$ is used in Eq.~\eqref{eq:f1}, and $B = \{0, 10, 20, 50\}$ in Eqs.~\eqref{eq:comp} and \eqref{eq:suff}. For the comparison analyses, we define overall feature importance based on the absolute value, and token removal is performed through masking with a designated mask token. Additionally, we conduct a qualitative comparison by reporting the top-$20\%$ most important tokens across selected dataset examples. This could form the basis for a future user study that provides a thorough qualitative evaluation of whether the methods yield explanations that are plausible to human users, beyond what is captured by quantitative metric scores.

For evaluating the method presented in Sec.~\ref{sec:attn_directional_derivative}, we adopt the experimental protocol used in prior work \citep{pmlr-v80-kim18d} to ensure comparability in the absence of standardised evaluation metrics for concept-based explanations. 
In the interest of ensuring consistency with the original study and not conflating method development with metric design, we view the development of novel evaluation metrics as an important but orthogonal line of inquiry, outside the scope of this work.

\section{Results and analysis}
\label{sec:results}

\subsection{Explaining transformer predictions for NLP classification}
\begin{figure}
    \centering
    \begin{subfigure}{\textwidth}\includegraphics[width=\textwidth]{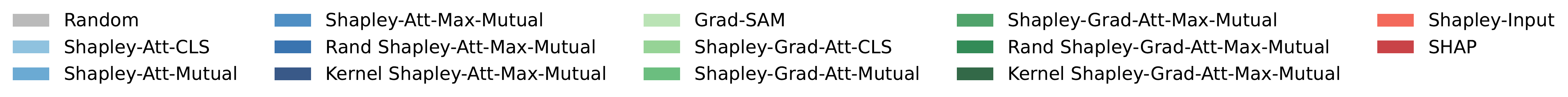}
    \end{subfigure}
    \\
    \begin{subfigure}{0.285\textwidth} 
        \includegraphics[width=\textwidth]{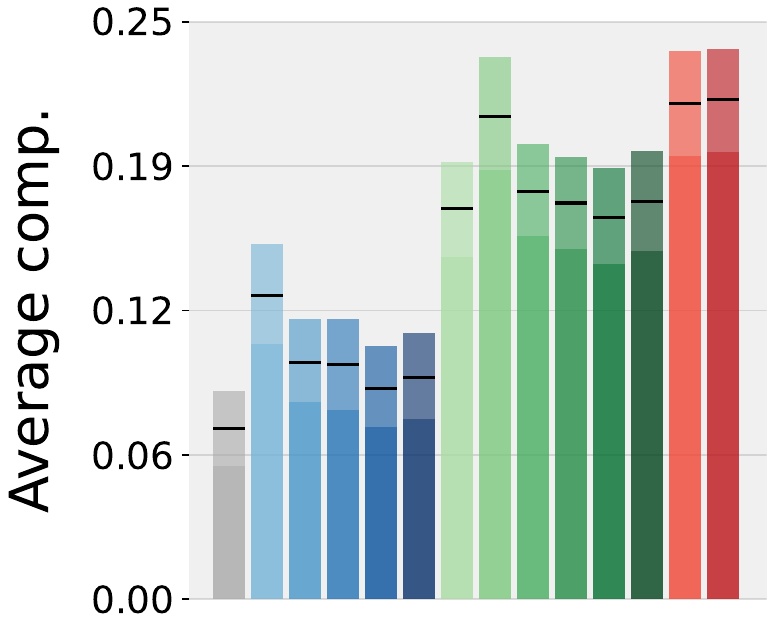}
        \caption{}
        \label{fig:sst2_short}
    \end{subfigure}
    \hfill
    \begin{subfigure}{0.233\textwidth} 
        \includegraphics[width=\textwidth]{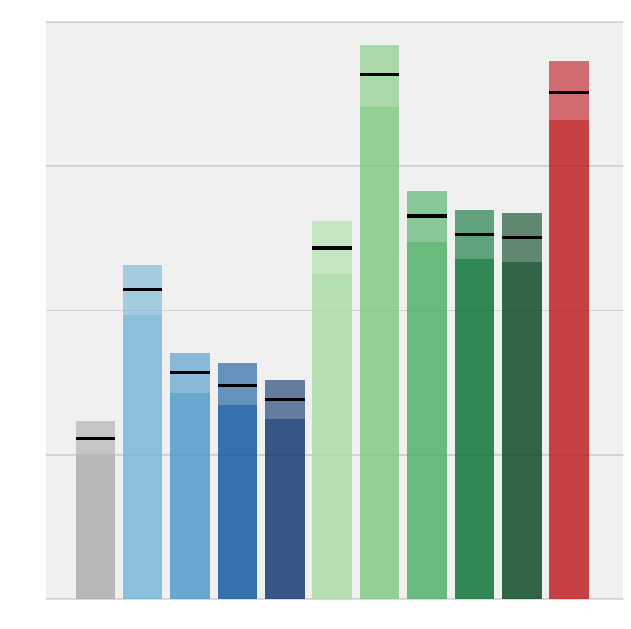}
        \caption{}
        \label{fig:sst2}
    \end{subfigure}
    \hfill
    \begin{subfigure}{0.233\textwidth}
        \includegraphics[width=\textwidth]{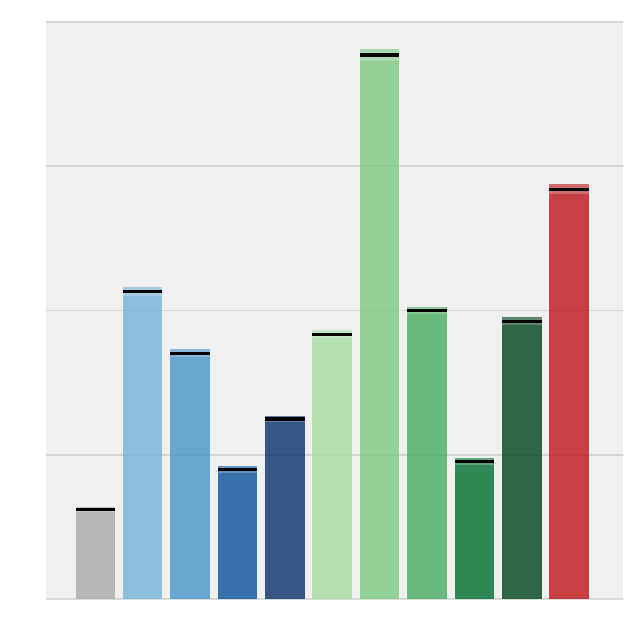}
        \caption{}
        \label{fig:imdb}
    \end{subfigure}
    \hfill
    \begin{subfigure}{0.233\textwidth}
        \includegraphics[width=\textwidth]{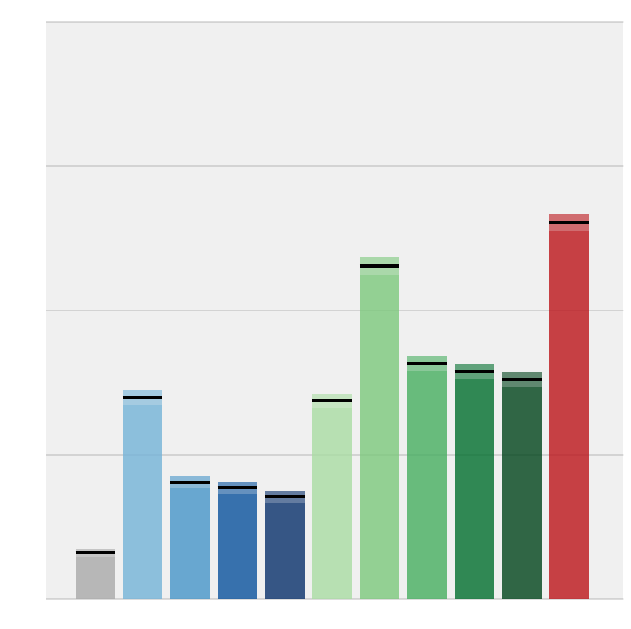}
        \caption{}
        \label{fig:ag_news}
    \end{subfigure}

    \caption{Average comprehensiveness (comp.) scores across all methods for datasets (a) SST-2 short, (b) SST-2, (c) IMDb, and (d) Ag News. The black horizontal lines mark the average values, and the lighter regions indicate the bootstrap confidence estimates. Methods are colour-coded such that blue corresponds to attention-based, green to gradient-based, and red to input-based methods.}
    \label{fig:nlp_barplot_comp}
\end{figure}

The performance of the explanation methods evaluated on the NLP task is shown in Fig.~\ref{fig:nlp_barplot_comp} and detailed in Tab.~\ref{tab:text_results}, and results for the remaining metrics from Sec.~\ref{sec:metrics} are provided in App.~\ref{app:npl_all_metrics}. We observe an overall agreement among the three metrics for each method. A qualitative comparison of selected methods is shown in Tab.~\ref{tab:nlp_qualitative_eval}.

The input-based methods, Shapley-Input and SHAP, consistently achieve strong performance across all evaluation metrics and datasets, followed by the gradient-based approaches. This former method category is expected to perform better as it leverages the full model. Under the formulations of the characteristic functions in Eqs.~\eqref{eq:char_func_grad_att_cls}, \eqref{eq:char_func_grad_att_mutual} and \eqref{eq:char_func_grad_att_mutual_max}, a key assumption is that attention weights are the main mediator of meaningful information to the model's final decision, refer to Sec.~\ref{sec:shaptention}. The empirical results indicate that there is not necessarily a correspondence between attributions from these methods and Shapley-Input for the same input token, a discrepancy that is likely due to the information lost when accepting this assumption. However, when accounting for uncertainty, the proposed Shapley-Grad-Att-CLS performs similarly to the SHAP framework, showing the effectiveness of using the classification token. Among the approximations, the kernel-based approach demonstrates the strongest results. It is expected that performance can be further improved by increasing the number of sampled coalitions or by employing more suitable sampling strategies, such as those suggested by \citet{olsen2024improving}. In contrast, the pure attention-based methods show notably weaker performance. This result is partly to be expected, as the selected metrics in Sec.~\ref{sec:metrics} are output-focused, measuring the ability to reconstruct or invert model predictions from masked input sequences. Attention weights themselves, aside from those associated with the classification token, are not necessarily directly indicative of the model's final output; rather, they encode broader linguistic phenomena concerning inter-token relations.

\definecolor{lightgray}{gray}{0.9}
\definecolor{blue}{rgb}{0.6, 0.8, 1}  
\definecolor{green}{rgb}{0.4, 0.8, 0.4}  
\definecolor{lightblue}{rgb}{0.8, 0.9, 1}  
\definecolor{lightgreen}{rgb}{0.75, 1, 0.75} 

\begin{table}[ht]
    \centering
    \renewcommand{\arraystretch}{1.5}
    \footnotesize
    \caption{Metric performance of all evaluated methods in Tab.~\ref{tab:methods_nlp} across all datasets. Uncertainties, estimated using lower and upper bounds of the 95\% bootstrap confidence interval, are stated for metrics comprehensiveness (Comp.) and sufficiency (Suff.). Bold values indicate the best results per metric, considering the uncertainty bounds. Methods are further grouped into attention-, gradient-, and input-based (refer to Fig.~\ref{fig:nlp_barplot_comp}). A dash ("-") denotes method-dataset combinations for which computing results is computationally infeasible.}
    \resizebox{\textwidth}{!}{
    \begin{tabular}{p{1.5cm} @{\hspace{20pt}}ccc@{\hspace{25pt}}ccc@{\hspace{25pt}}ccc@{\hspace{25pt}}ccc}
        \toprule
        \toprule
         & \multicolumn{3}{c}{\hspace{-30pt} \textbf{SST-2 short}} 
         & \multicolumn{3}{c}{\hspace{-20pt} \textbf{SST-2}} 
         & \multicolumn{3}{c}{\hspace{-35pt} \textbf{IMDb}} 
         & \multicolumn{3}{c}{\hspace{-20pt} \textbf{Ag News}} \\
        \textbf{Methods} & \textbf{F1} $\uparrow$ & \textbf{Comp.} $\uparrow$ & \textbf{Suff.} $\downarrow$ 
        & \textbf{F1} & \textbf{Comp.} & \textbf{Suff.}  
        & \textbf{F1} & \textbf{Comp.} & \textbf{Suff.}  
        & \textbf{F1} & \textbf{Comp.} & \textbf{Suff.}  \\
        \midrule
        \midrule
        Random & 0.627 & 0.074$^{\scriptsize +0.016}_{\scriptsize -0.015}$ & 0.267$ {\scriptsize\pm0.024}$ & 0.630 & 0.070$ {\scriptsize\pm0.008}$ & 0.279$^{\scriptsize +0.015}_{\scriptsize -0.016}$ & 0.570 & 0.039$ {\scriptsize\pm0.001}$ & 0.302$ {\scriptsize\pm0.002}$ & 0.746 & 0.020$ {\scriptsize\pm0.002}$ & 0.296$ {\scriptsize\pm0.004}$ \\
        \midrule
        Shapley-Att-CLS & 0.791 & 0.132$ {\scriptsize\pm0.022}$ & 0.210$^{\scriptsize +0.022}_{\scriptsize -0.020}$ & 0.735 & 0.134$ {\scriptsize\pm0.011}$ & 0.221$ {\scriptsize\pm0.015}$ & 0.752 & 0.133$ {\scriptsize\pm0.002}$ & 0.237$ {\scriptsize\pm0.002}$ & 0.897 & 0.087$ {\scriptsize\pm0.003}$ & 0.208$ {\scriptsize\pm0.003}$ \\
        Shapley-Att-Mutual & 0.688 & 0.102$^{\scriptsize +0.019}_{\scriptsize -0.018}$ & 0.256$ {\scriptsize\pm0.025}$ & 0.667 & 0.098$ {\scriptsize\pm0.009}$ & 0.257$ {\scriptsize\pm0.015}$ & 0.779 & 0.106$ {\scriptsize\pm0.002}$ & 0.242$ {\scriptsize\pm0.002}$ & 0.827 & 0.051$ {\scriptsize\pm0.003}$ & 0.245$ {\scriptsize\pm0.004}$ \\
        Shapley-Att-Max-Mutual & 0.698 & 0.102$^{\scriptsize +0.020}_{\scriptsize -0.018}$ & 0.251$ {\scriptsize\pm0.024}$ & - & - & - & - & - & - & - & - & - \\
        Rand Shapley-Att-Max-Mutual & 0.685 & 0.091$ {\scriptsize\pm0.017}$ & 0.270$^{\scriptsize +0.025}_{\scriptsize -0.024}$ & 0.633 & 0.093$ {\scriptsize\pm0.009}$ & 0.257$ {\scriptsize\pm0.014}$ & 0.566 & 0.056$ {\scriptsize\pm0.001}$ & 0.294$ {\scriptsize\pm0.002}$ & 0.825 & 0.048$ {\scriptsize\pm0.003}$ & 0.244$ {\scriptsize\pm0.004}$ \\
        Kernel Shapley-Att-Max-Mutual & 0.653 & 0.096$ {\scriptsize\pm0.019}$ & 0.258$^{\scriptsize +0.025}_{\scriptsize -0.023}$ & 0.653 & 0.086$ {\scriptsize\pm0.008}$ & 0.265$ {\scriptsize\pm0.016}$ & 0.658 & 0.078$ {\scriptsize\pm0.001}$ & 0.273$ {\scriptsize\pm0.002}$ & 0.834 & 0.044$ {\scriptsize\pm0.003}$ & 0.246$ {\scriptsize\pm0.004}$ \\
        \midrule
        Grad-SAM & 0.859 & 0.169$ {\scriptsize\pm0.021}$ & \textbf{0.166}$ {\scriptsize\pm0.019}$ & 0.804 & 0.152$ {\scriptsize\pm0.010}$ & 0.191$ {\scriptsize\pm0.012}$ & 0.782 & 0.115$ {\scriptsize\pm0.002}$ & 0.229$ {\scriptsize\pm0.002}$ & 0.895 & 0.086$ {\scriptsize\pm0.003}$ & 0.201$ {\scriptsize\pm0.003}$ \\
        Shapley-Grad-Att-CLS & 0.873 & \textbf{0.209}$ {\scriptsize\pm0.025}$ & \textbf{0.137}$ {\scriptsize\pm0.017}$ & 0.845 & \textbf{0.227}$ {\scriptsize\pm0.013}$ & \textbf{0.153}$ {\scriptsize\pm0.012}$ & 0.812 & \textbf{0.236}$ {\scriptsize\pm0.002}$ & 0.208$ {\scriptsize\pm0.002}$ & 0.928 & 0.144$ {\scriptsize\pm0.004}$ & 0.170$ {\scriptsize\pm0.002}$ \\
        Shapley-Grad-Att-Mutual & 0.877 & \textbf{0.176}$ {\scriptsize\pm0.021}$ & \textbf{0.160}$ {\scriptsize\pm0.018}$ & 0.823 & 0.166$ {\scriptsize\pm0.011}$ & 0.178$ {\scriptsize\pm0.012}$ & 0.810 & 0.125$ {\scriptsize\pm0.002}$ & 0.213$ {\scriptsize\pm0.002}$ & 0.911 & 0.102$ {\scriptsize\pm0.003}$ & 0.187$ {\scriptsize\pm0.003}$ \\
        Shapley-Grad-Att-Max-Mutual & \textbf{0.886} & 0.172$ {\scriptsize\pm0.020}$ & \textbf{0.166}$^{\scriptsize +0.018}_{\scriptsize -0.017}$ & - & - & - & - & - & - & - & - & - \\
        Rand Shapley-Grad-Att-Max-Mutual & 0.859 & 0.165$ {\scriptsize\pm0.021}$ & \textbf{0.169}$^{\scriptsize +0.018}_{\scriptsize -0.017}$ & 0.802 & 0.158$ {\scriptsize\pm0.011}$ & 0.187$ {\scriptsize\pm0.012}$ & 0.569 & 0.059$ {\scriptsize\pm0.001}$ & 0.287$ {\scriptsize\pm0.003}$ & 0.911 & 0.099$ {\scriptsize\pm0.003}$ & 0.188$ {\scriptsize\pm0.003}$ \\
        Kernel Shapley-Grad-Att-Max-Mutual & \textbf{0.886} & \textbf{0.172}$^{\scriptsize +0.023}_{\scriptsize -0.022}$ & \textbf{0.164}$ {\scriptsize\pm0.018}$ & 0.799 & 0.157$ {\scriptsize\pm0.011}$ & 0.186$ {\scriptsize\pm0.012}$ & 0.804 & 0.120$ {\scriptsize\pm0.002}$ & 0.218$ {\scriptsize\pm0.002}$ & 0.912 & 0.095$ {\scriptsize\pm0.003}$ & 0.189$ {\scriptsize\pm0.003}$ \\
        \midrule
        Shapley-Input & \textbf{0.886} & \textbf{0.215}$^{\scriptsize +0.023}_{\scriptsize -0.021}$ & \textbf{0.145}$^{\scriptsize +0.015}_{\scriptsize -0.014}$ & - & - & - & - & - & - & - & - & - \\
        SHAP & \textbf{0.886} & \textbf{0.216}$ {\scriptsize\pm0.023}$ & \textbf{0.138}$^{\scriptsize +0.017}_{\scriptsize -0.016}$ & \textbf{0.873} & \textbf{0.219}$ {\scriptsize\pm0.012}$ & \textbf{0.149}$ {\scriptsize\pm0.009}$ & \textbf{0.859} & 0.178$ {\scriptsize\pm0.002}$ & \textbf{0.179}$ {\scriptsize\pm0.002}$ & \textbf{0.936} & \textbf{0.163}$ {\scriptsize\pm0.004}$ & \textbf{0.160}$ {\scriptsize\pm0.002}$ \\
\bottomrule
        \bottomrule
    \end{tabular}
    }
    \label{tab:text_results}
\end{table}

In addition to being competitive with SHAP in terms of metric performance, the attention-based Shapley values from Sec.~\ref{sec:shaptention} offer notable advantages in computational efficiency. Specifically, the computation requires only a single forward pass to extract attention weights from all layers and heads, followed by a backward pass to compute the associated gradients. In contrast, both exact Shapley value and SHAP computations involve masking input features over several iterations and performing multiple forward passes to obtain corresponding model predictions, leading to substantially higher computational demands. As such, attention-based Shapley values present a promising alternative, balancing the commonly cited benefit of its theoretical foundation with practical efficiency.  

To further explore the relation between the attention mechanism and Shapley values in the context of feature-level interactions, we examine Shapley interaction values, which quantify the joint contribution of feature pairs to the model's output. \citet{lundberg2020local} propose \textit{SHAP interaction values} describing local interaction effects based on the Shapley interaction index from cooperative game theory. Following this formulation, we use exact Shapley values similar to Shapley-Input from Tab.~\ref{tab:methods_nlp}. In preliminary analysis of selected sample sentences, we observe weak and inconsistent correlations between raw attention weights and Shapley interaction values. In rather using the gradient of the model output with respect to the attention weights, we find stronger correlations across more layers and heads. Nonetheless, attention mechanisms and Shapley-based frameworks embody fundamentally different perspectives of model explanation. While Shapley interaction values explicitly quantify the joint feature contribution to the model’s output, attention weights serve as a structural mechanism that governs how information is shared between tokens. 

\subsection{Explaining transformer concept sensitivity for image classification}

\begin{figure}
    \centering

    \begin{subfigure}{\textwidth}
        \includegraphics[width=\textwidth]{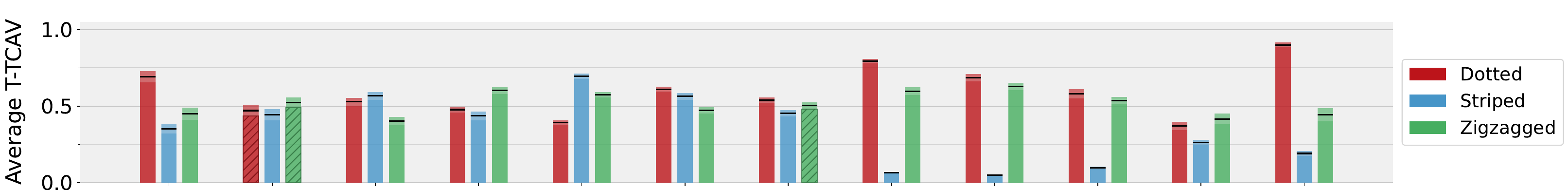}
        \caption{}
    \end{subfigure}
    \vfill

    \begin{subfigure}{\textwidth}
        \includegraphics[width=\textwidth]{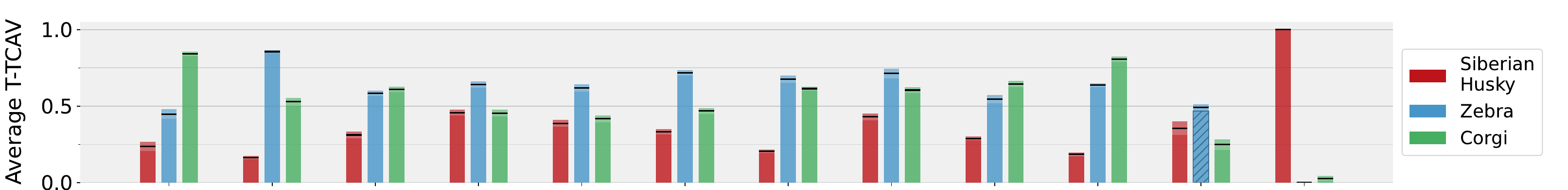}
        \caption{}
    \end{subfigure}
    \vfill  

    \begin{subfigure}{\textwidth}
        \includegraphics[width=\textwidth]{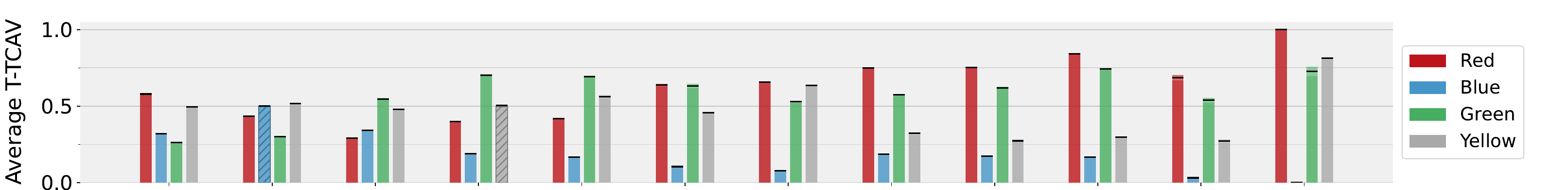}
        \caption{}
    \end{subfigure}
    \vfill

    \begin{subfigure}{\textwidth}
        \includegraphics[width=\textwidth]{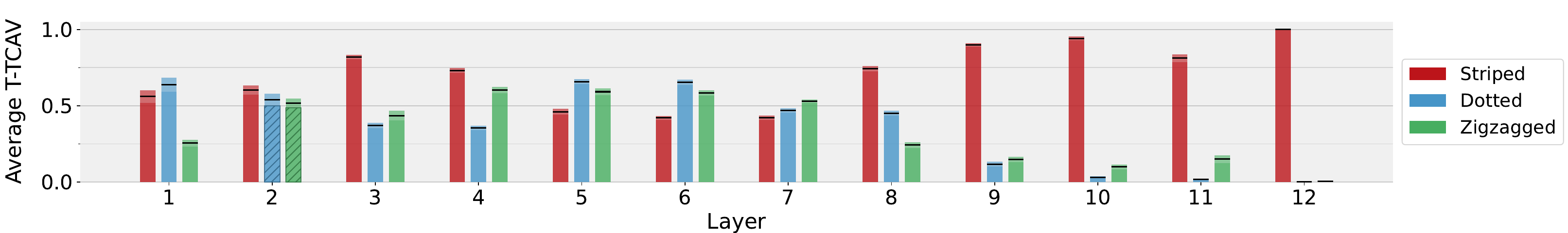}
        \caption{}
    \end{subfigure}

    \caption{The average T-TCAV scores computed across 50 relative CAVs for the target classes (a) \textit{dalmatian}, (b) \textit{dog sled}, (c) \textit{fire engine}, and (d) \textit{zebra}, evaluated for various concepts and reported for all encoder layers. For each concept, the negative dataset is formed by combining samples from all other concepts, refer to Sec.~\ref{sec:datasets}. The lighter bar regions indicate confidence intervals, and bars with a hatched pattern mark scores that are not statistically significant from a two-sided t-test, under the null hypothesis that the true T-TCAV score equals 0.5.}
    \label{fig:vit_barplot}
\end{figure}

Results for the proposed T-TCAV method from Sec.~\ref{sec:attn_directional_derivative} are presented in Fig.~\ref{fig:vit_barplot}, computed for all ViT encoder layers. Specifically, these results show the average T-TCAV scores computed across 50 relative CAVs, each trained using different randomly sampled negative sets. Following \citet{pmlr-v80-kim18d}, statistical significance is assessed by conducting a two-sided t-test on all T-TCAV scores obtained from these samples. The null hypothesis represents the assumption that there is no association between the concept and the target class, i.e., the true T-TCAV score is 0.5. If the null hypothesis is rejected (i.e., $p < 0.05$), the concept is interpreted as being statistically significant to the model's class prediction.

In neural networks, deeper layers typically represent increasingly abstract and high-level features \citep{pmlr-v80-kim18d, 10.1007/978-3-319-10590-1_53}, and concepts most aligned with the target class are expected to be more strongly represented near the output. As shown in Fig.~\ref{fig:vit_barplot}, in most cases, the concept presumed to be most semantically aligned with the target class tends to dominate in the deeper layers. This suggests that the model increasingly relies on the most relevant concepts in the final stages of computation. The observed variability in Fig.~\ref{fig:vit_barplot} is likely caused by the instability of CAVs, as these are sensitive to factors such as the composition and representativeness of the concept dataset, as well as the architecture and training regime of the probe classifier. 

Figure~\ref{fig:tcav_vs_ttcav} shows a direct comparison of the T-TCAV score with the original TCAV method by \citet{pmlr-v80-kim18d}, under identical experimental conditions. As expected, we observe differences in the results between the two approaches. Although it is evident that attention weights contribute additional information, neither score can be claimed to provide a definitive ground truth for global concept sensitivity. Instead, T-TCAV combines two different forms of information about the model, namely attention weights and concept sensitivity, while TCAV offers merely the latter. While we argue that T-TCAV yields a more holistic picture in the transformer context, it remains the responsibility of the practitioner to determine whether this additional information is relevant for the XAI question at hand.

\begin{figure}[h]
    \centering
    
    \begin{subfigure}[b]{0.19\textwidth}
        \includegraphics[width=\textwidth]{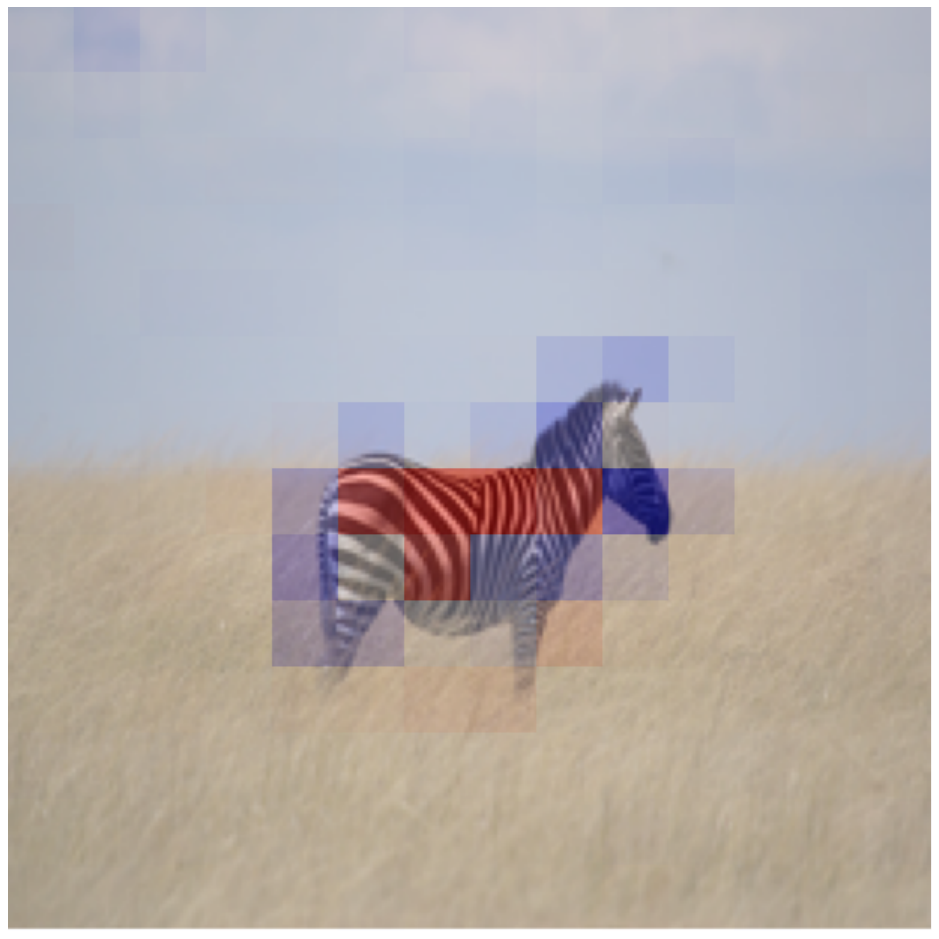}
        \caption{}
        \label{fig:zebra_1}
    \end{subfigure}
    \hfill
    \begin{subfigure}[b]{0.19\textwidth}
        \includegraphics[width=\textwidth]{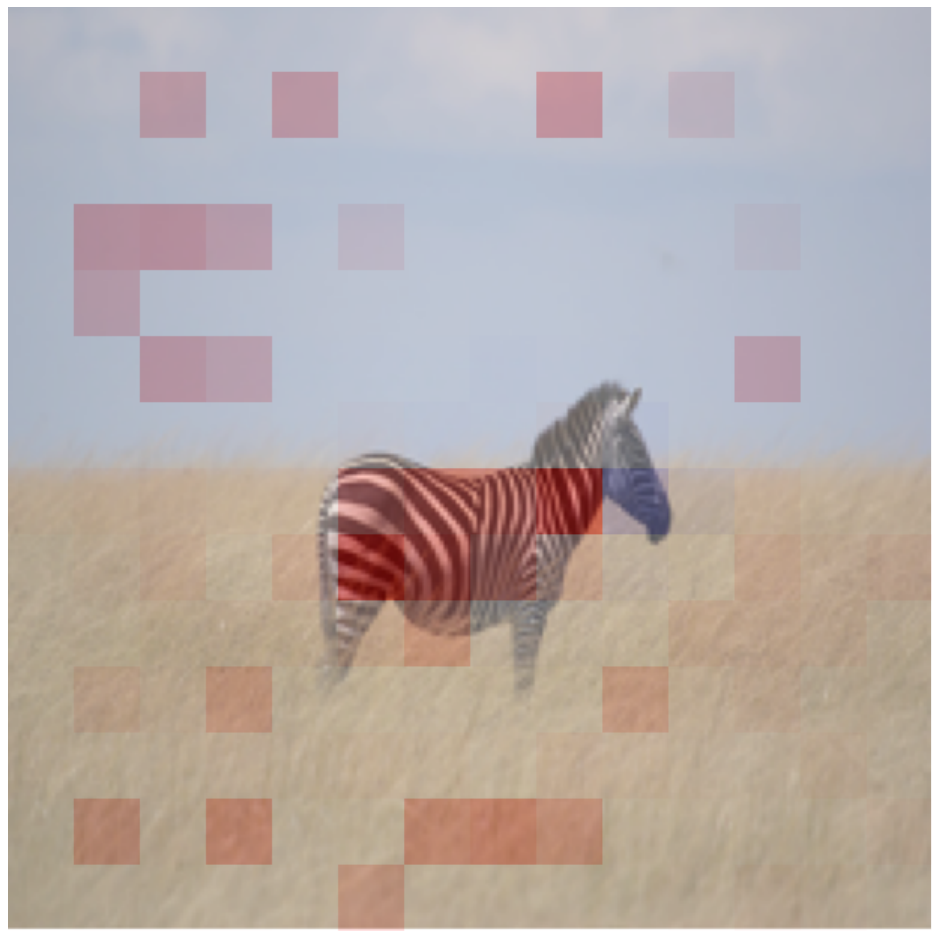}
        \caption{}
        \label{fig:zebra_2}
    \end{subfigure}
    \hfill
    \begin{subfigure}[b]{0.19\textwidth}
        \includegraphics[width=\textwidth]{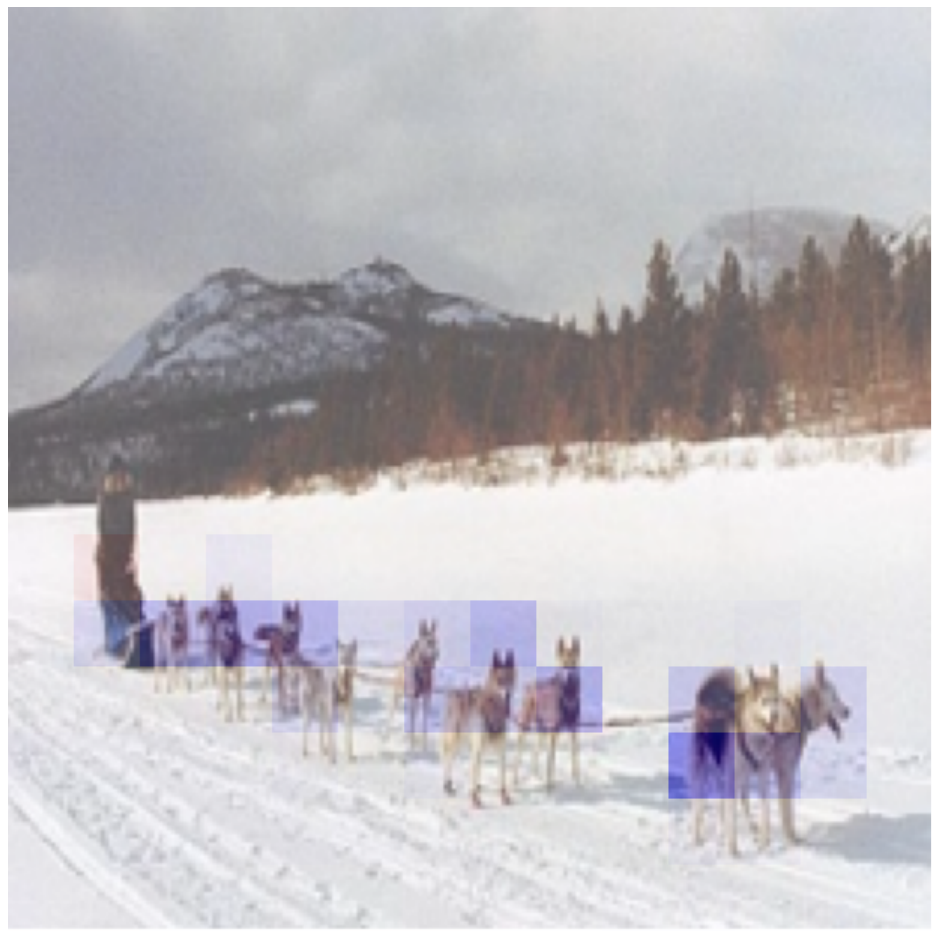}
        \caption{}
        \label{fig:dog_sled}
    \end{subfigure}
    \hfill
    \begin{subfigure}[b]{0.19\textwidth}
        \includegraphics[width=\textwidth]{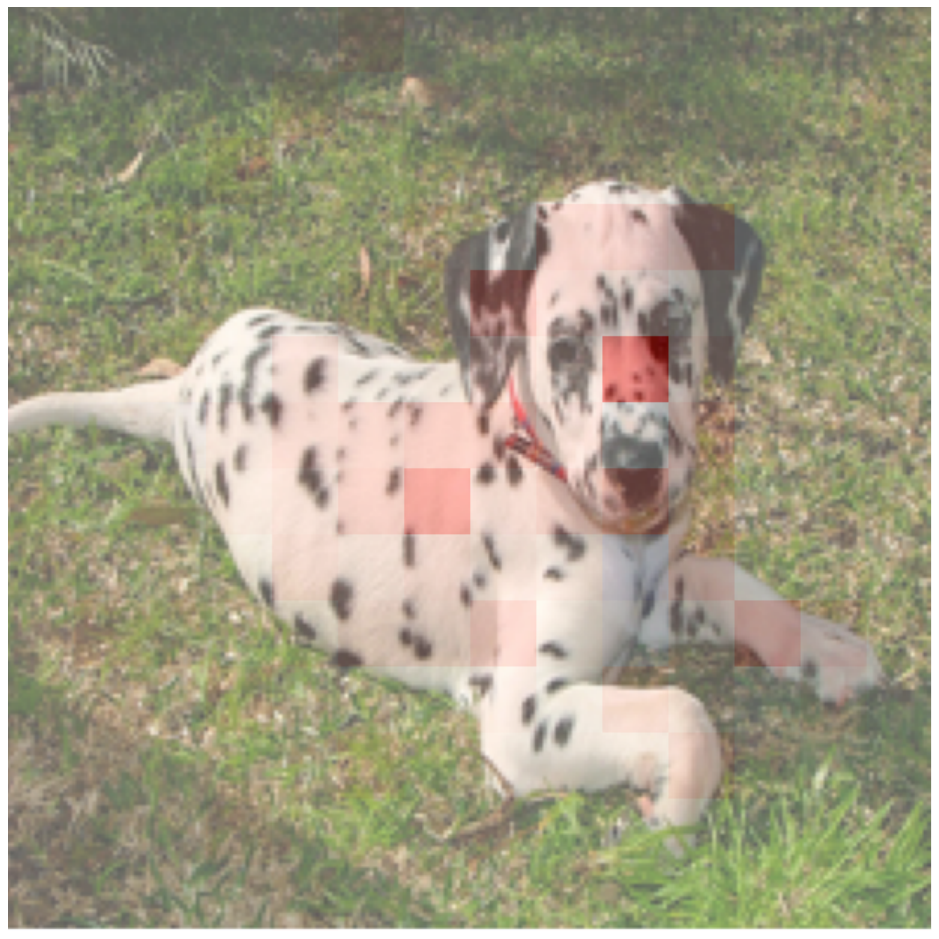}
        \caption{}
        \label{fig:dalmatian_1}
    \end{subfigure}
    \hfill
    \begin{subfigure}[b]{0.19\textwidth}
        \includegraphics[width=\textwidth]{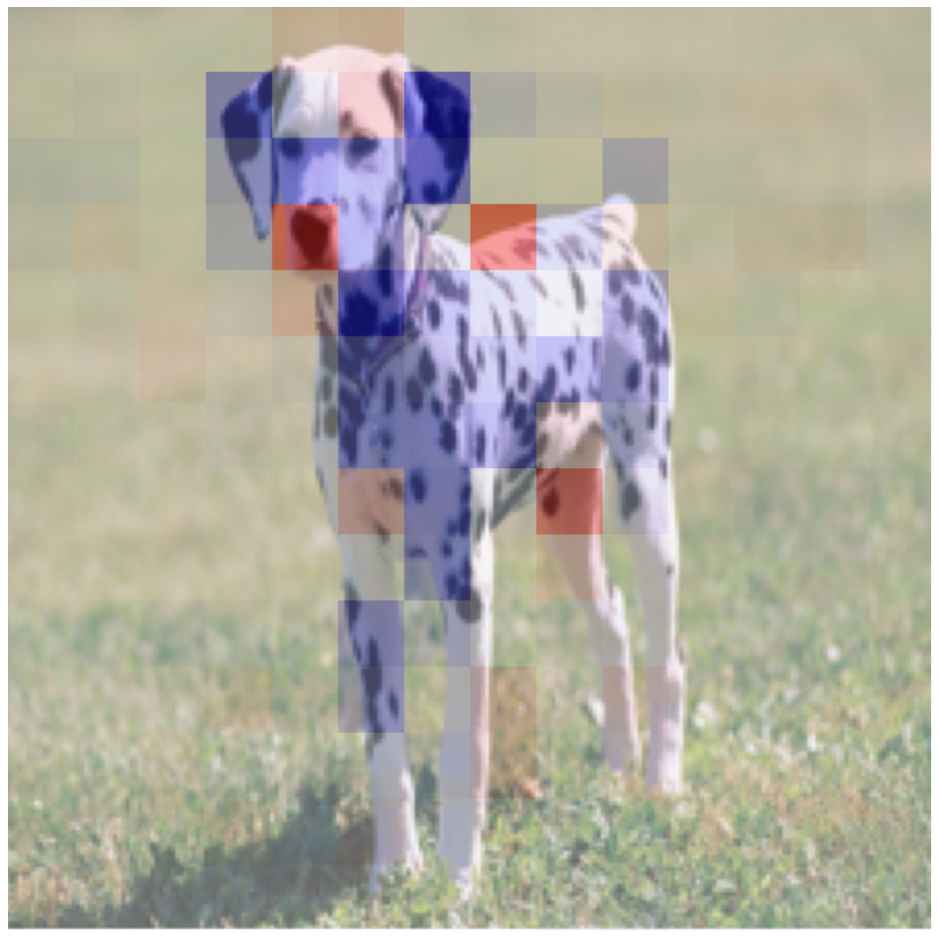}
        \caption{}
        \label{fig:dalmatian_2}
    \end{subfigure}

    \caption{Heatmap overlays of the attention-weighted directional derivatives defined in Eq.~\eqref{eq:token_directional_derivative} for selected concepts at specific layers: (a) \textit{striped} at layer 5; (b) \textit{striped} at layer 10; (c) \textit{Siberian Husky} at layer 9; (d, e) \textit{dotted} at layer 6. All images are retrieved from ImageNet \citep{5206848}. Colour scales are relative per image.}
    \label{fig:vit_heatmap}
\end{figure}

The token-level, attention-weighted directional derivatives defined in Eq.~\eqref{eq:token_directional_derivative} may serve as a tool for generating fine-grained local explanations by identifying input regions that are more sensitive to a given concept. In Fig. \ref{fig:zebra_1},~b, we visualise heatmap overlays of the evolution of concept sensitivity for the \textit{zebra} target class in relation to the \textit{stripes} concept. In the deeper layers, more conceptually relevant regions are detected with positive directional derivatives, while also increasingly highlighting surrounding regions. Fig. \ref{fig:dog_sled} further illustrates that, for the \textit{Siberian Husky} concept in the dog sled scene, the relevant regions are predominantly associated with strong negative directional derivatives. This is consistent with the lower T-TCAV scores observed for the same target class in Fig.~\ref{fig:vit_barplot}, suggesting the concept is less positively aligned with the model's decision in this context. We hypothesise that this is due to one of two factors: either, the information as represented by the CAV is insufficient for the classification task, or the CAV does not represent the intended concept but rather a proxy that correlates strongly with the concept dataset.
Figs. \ref{fig:dalmatian_1},~e show two different images from the same target class, visualised at the same layer. In the former, the regions of the \textit{dalmatian} image that visibly exhibit the \textit{dotted} concept are associated with positive directional derivatives. By contrast, in Fig. \ref{fig:dalmatian_2}, those same relevant regions are partly associated with negative directional derivatives, illustrating an inconsistency in how concept-relevant regions are attributed across inputs within the same target class and model layer. 

\section{Discussion and conclusion}
\label{sec:discussion_conclusion}

This study integrates attention weights into well-established XAI frameworks for producing explanations of transformer models. We introduce two novel explanation methods suitable for both NLP and image classification tasks: one formulating a cooperative game with an attention-based characteristic function as the basis for a Shapley decomposition, while the other incorporates attention weights into token-level directional derivatives for computing concept sensitivity. The attention weights in a transformer model do not directly determine the model's output, but influence it indirectly, offering complementary insight into the model's operations. The proposed methods have been designed with the aim of integrating precisely this insight into model explanations.

The two proposed methods are both, in principle, directly applicable to text and image data, although we do not show empirical results for both modalities in this study. Their applicability to both modalities is given by the fact that the methods have no modality-specific assumptions or implementation constraints. However, there are challenges associated with the application of the proposed methods to the respective data modalities, which we briefly describe in the following. In the case of image data, only approximations of the attention-based Shapley values with, e.g., random or kernel-based sampling of coalitions, are feasible to obtain, as the complexity of the Shapley decomposition scales exponentially with the number of players, corresponding to the number of image patches. In the case of text data, the application of any concept-based method (including TCAV and T-TCAV) requires access to concept datasets appropriate for the relevant domain in order to construct meaningful CAVs. Constructing high-quality concept datasets can be significantly resource-demanding.

Although the proposed attention-based Shapley decompositions are understood as attribution methods\footnote{The attribution aspect of our method comes primarily from raw model gradients. Other attribution frameworks, such as Integrated Gradients \citep{sundararajan2017axiomatic}, may be adapted to incorporate attention weights and constitute a potential direction for further efforts in attention-based model explanations.}, the aspects of model behaviour captured by attention weights are not necessarily reflected in output-focused evaluation criteria of standard quantitative XAI metrics. Evaluating such aspects constitutes a complementary but distinct form of assessment. To the best of our knowledge, no widely accepted metric exists for quantifying the extent to which an explanation reflects internal representational dynamics. One possible approach would be to directly intervene on the attention weights by randomising or adversarially modifying their distributions. However, attention weights cannot be meaningfully manipulated in isolation, as they are learned jointly with other parameters that together determine their functional role. Perturbing attention weights without re-training disrupts this learned coupling and produces a model state that does not correspond to any valid trained solution \citep{wiegreffe-pinter-2019-attention}. This highlights a broader challenge in XAI: current metrics may not fully capture the value of explanations that reflect internal model dynamics rather than direct input-output influence. However, TCAV can be interpreted as a metric for concept activation vectors, as it yields a single number quantifying how consistently a concept direction in the model's internal representation space influences the prediction. If adopting this interpretation of TCAV as a metric, our method of T-TCAV should also be interpreted as a metric, but one that provides additional information about the internal model representation, in the form of attention weights. Developing general quantitative evaluation metrics that incorporate such explanation aspects remains a subject beyond the scope of the present work. Also relevant yet beyond the present scope is the use of structured human evaluations to assess the extent to which attention-based explanations align with human reasoning. 

Both the definition of the characteristic function in the game formulation for the Shapley decomposition and the nature of latent concept representations in transformer models influence the effectiveness of the corresponding explanation method. A single approach to transformer XAI is likely insufficient, highlighting the benefit of the versatility and flexibility of attention weights for explanation purposes.

\bibliographystyle{plainnat}  
\bibliography{references}  

\clearpage

\appendix

\renewcommand{\thesection}{\Alph{section}}

\section{Characteristic functions for the Shapley decomposition}
\label{app:characteritic_functions}

Sections \ref{app:char_func_shapley_input} and \ref{app:char_func_shapley_att} define the characteristic function for methods Shapley-Input and Shapley-Att, respectively (see Tab.~\ref{tab:methods_nlp}), used in computing Shapley values according to Eq.~\eqref{eq:shapley_values}. Section \ref{app:derivation_shapley_mutual_attn} provides the expression of the Shapley value based on mutual attention interactions.

\subsection{Characteristic function for Shapley-Input}
\label{app:char_func_shapley_input}
Given a model input sequence $x$ including special tokens, let $x_P$ denote the subsequence of original (non-special) tokens with a corresponding index set $\mathcal{X}_P$. For any coalition $\mathcal{S} \subseteq \mathcal{X}_P$, a modified input sequence $x_{\mathcal{S}}$ is constructed by replacing all tokens in $x_P$ at positions not in $\mathcal{S}$ with a designated masking token, while keeping the special tokens fixed. The value assigned to coalition $\mathcal{S}$ is defined as 

\begin{equation}
    v(\mathcal{S}) = f_k(x_{\mathcal{S}}),
    \label{eq:char_func_shapley_input}
\end{equation}

where $f_k$ denotes the model prediction for class $k$ and $v(\emptyset) = 0$. 

\subsection{Characteristic functions for attention-based Shapley decomposition}
\label{app:char_func_shapley_att}

The average attention matrix across all layers and heads is defined as

\begin{equation}
    \mathbf{A} = \frac{1}{LH} \sum_{\ell = 1}^{L} \sum_{h = 1}^H \mathbf{A}^{\ell h},
    \label{eq:avg_attn}
\end{equation}

where $\mathbf{A}^{\ell h}$ denotes the attention matrix corresponding to layer $\ell \in \{1, ..., L\}$ and head $h \in \{1, ..., H\}$.

Equivalent characteristic functions to Eqs.~\eqref{eq:char_func_grad_att_cls}, \eqref{eq:char_func_grad_att_mutual} and \eqref{eq:char_func_grad_att_mutual_max} when omitting gradient information can be obtained based on the averaged matrix $\mathbf{A}$ in Eq.~\eqref{eq:avg_attn}. Following the notation introduced in Sec.~\ref{sec:shaptention}, the value of a coalition $\mathcal{S} \subseteq \mathcal{X}_P$ for methods Shapley-Att-CLS, Shapley-Att-Mutual, and Shapley-Att-Max-Mutual (see Tab.~\ref{tab:methods_nlp}) are

\begin{equation}
    v(\mathcal{S}) =     
    \begin{cases}
      \sum\limits_{i \in \mathcal{S}} A_{\text{CLS}, i}, & \text{if } |\mathcal{S}| \geq 1 \\
      0, & \text{otherwise} \\
    \end{cases},
    \label{eq:char_func_att_cls}
\end{equation}

\begin{equation}
    v(\mathcal{S}) =     
    \begin{cases}
      \sum\limits_{\substack{i, j \in \mathcal{S} \\ i \neq j}} \left( A_{ij} + A_{ji} \right), & \text{if } |\mathcal{S}| > 1 \\
      A_{ii}, & \text{if } \mathcal{S} = \{i\} \\
      0, & \text{otherwise} \\
    \end{cases},
    \label{eq:char_func_att_mutual}
\end{equation}

and

\begin{equation}
    v(\mathcal{S}) =     
    \begin{cases}
      \sum\limits_{\substack{i, j \in \mathcal{S} \\ i \neq j}} \text{max}\left( A_{ij}, \, A_{ji} \right), & \text{if } |\mathcal{S}| > 1 \\
      A_{ii}, & \text{if } \mathcal{S} = \{i\} \\
      0, & \text{otherwise} \\
    \end{cases},
    \label{eq:char_func_att_mutual_max}
\end{equation}

respectively.

\subsection{Shapley values based on mutual attention interactions}
\label{app:derivation_shapley_mutual_attn}
The Shapley decomposition formula in Eq.~\eqref{eq:shapley_values} calculated using either of the characteristic functions defined in Eqs.~\eqref{eq:char_func_grad_att_mutual} or \eqref{eq:char_func_att_mutual} can be reformulated to avoid having to compute the sum over all possible coalitions. This section presents the derivation for Shapley-Grad-Att-Mutual using the characteristic function in Eq.~\eqref{eq:char_func_grad_att_mutual}; see Tab.~\ref{tab:methods_nlp} for reference.   

The Shapley value for player $i$, $\phi_i$ in Eq.~\eqref{eq:shapley_values}, can be split into three sums, namely the empty coalition, all coalitions constituting a single player, and all remaining coalitions, hereafter referred to as Case 1, Case 2, and Case 3, respectively. Given a player set $\mathcal{N} = \{1, ..., N\}$ and the characteristic function stated in Eq.~\eqref{eq:char_func_grad_att_mutual}, the three sums following the three defined cases are derived as follows:

\paragraph{Case 1}
The part of $\phi_i$ including the empty coalition reduces to
\begin{equation}
    \sum_{\mathcal{S} = \emptyset} \frac{|\mathcal{S}|! (N - |\mathcal{S}| - 1)!}{N!} \left( v(\mathcal{S} \cup \{i\}) - v(\mathcal{S}) \right) = \frac{(N-1)!}{N!} v(\{i\}) = \frac{1}{N}M_{k, ii}.
\end{equation}

\paragraph{Case 2} 
The part of $\phi_i$ including all coalitions constituting a single player $j$ is

\begin{align}
    \sum\limits_{\substack{\mathcal{S} = \{j\} \\ i \neq j}} \frac{|\mathcal{S}|! (N - |\mathcal{S}| - 1)!}{N!}
    &\left( v(\mathcal{S} \cup \{i\}) - v(\mathcal{S}) \right) \nonumber \\
    &= \sum_{i \neq j} \frac{(N-2)!}{N!} \left(v(\{i, j\}) - v(\{j\}) \right) \nonumber \\
    &= \sum_{i \neq j} \frac{1}{N(N-1)}(M_{k, ij} + M_{k, ji} - M_{k, jj}).
\end{align}

\paragraph{Case 3}
The part of $\phi_i$ including remaining coalitions is computed as follows. First, the value of the characteristic function for a coalition $\mathcal{S}$ when including player $i$ is expressed as 

\begin{equation}
    v(\mathcal{S} \cup \{i\}) = v(\mathcal{S}) + \sum_{j \in \mathcal{S}} (M_{k, ij} + M_{k, ji}).
\end{equation}

Then, this relation replaces the marginal contribution of player $i$ to coalition $\mathcal{S}$ in the Shapley value decomposition, 

\begin{align}
    \sum\limits_{\substack{\mathcal{S} \subseteq \mathcal{N} \setminus \{i\} \\ |\mathcal{S}| \geq 2}} 
    \frac{|\mathcal{S}|! (N - |\mathcal{S}| - 1)!}{N!} 
    \left( v(\mathcal{S} \cup \{i\}) - v(\mathcal{S}) \right)
    &= 
    \sum\limits_{\substack{\mathcal{S} \subseteq \mathcal{N} \setminus \{i\} \\ |\mathcal{S}| \geq 2}} 
    \frac{|\mathcal{S}|! (N - |\mathcal{S}| - 1)!}{N!} 
    \sum_{j \in \mathcal{S}} (M_{k, ij} + M_{k, ji}) \nonumber \\
    &=
    \sum_{i \neq j} (M_{k, ij} + M_{k, ji}) 
    \sum\limits_{\substack{\mathcal{S} \subseteq \mathcal{N} \setminus \{i\} \\ |\mathcal{S}| \geq 2 \\ j \in \mathcal{S}}} 
    \frac{|\mathcal{S}|! (N - |\mathcal{S}| - 1)!}{N!} \nonumber \\
    &=
    \sum_{i \neq j} (M_{k, ij} + M_{k, ji}) 
    \sum_{z=2}^{N-1} \binom{N-2}{z-1} \frac{z! (N - z - 1)!}{N!},
\end{align}
where $z = |\mathcal{S}| \in \{2, ..., N - 1\}$ denotes the coalition size.
The coefficient term can be further simplified as
\begin{align}
\sum_{z=2}^{N-1} \binom{N-2}{z-1}
\frac{z! (N - z - 1)!}{N!}
&= \sum_{z=2}^{N-1}
\frac{(N - 2)!}{(z-1)! (N-z-1)!}
\frac{z! (N - z - 1)!}{N!} \nonumber  \\
&= \sum_{z=2}^{N-1}
\frac{z! (N-2)!}{(z-1)!N!}
\nonumber  \\
&= \sum_{z=2}^{N-1}
\frac{(N-2)!}{N!}\, z \nonumber \\
&= \frac{1}{N(N-1)}
\sum_{z=2}^{N-1} z
\nonumber  \\
&= \frac{1}{N(N-1)}
\left( \frac{N(N-1)}{2} - 1 \right) \nonumber \\
&= \frac{1}{2} - \frac{1}{N(N-1)} .
\end{align}
The combined expression yielding the final Shapley value for player $i$ is then

\begin{align}
    \phi_i &= \frac{1}{N}M_{k, ii} + \sum_{i \neq j} \frac{1}{N(N-1)}(M_{k, ij} + M_{k, ji} - M_{k, jj}) + \sum_{i \neq j} \left(\frac{1}{2} - \frac{1}{N(N-1)}\right) (M_{k, ij} + M_{k, ji}) \nonumber \\
    &=
    \frac{1}{N}M_{k, ii} + \frac{1}{N(N-1)} \sum_{i \neq j}(M_{k, ij} + M_{k, ji} - M_{k, jj}) + \left( \frac{1}{2} - \frac{1}{N(N-1)} \right) \sum_{i \neq j} (M_{k, ij} + M_{k, ji}) . 
\end{align}

\section{Additional results of explanation method performance on the NLP classification task}
\label{app:npl_all_metrics}

\FloatBarrier

\begin{figure}[H]
    \centering
    \begin{subfigure}{\textwidth}\includegraphics[width=\textwidth]{Fig1a.pdf}
    \end{subfigure}
    \\
    \begin{subfigure}{0.280\textwidth} 
        \includegraphics[width=\textwidth]{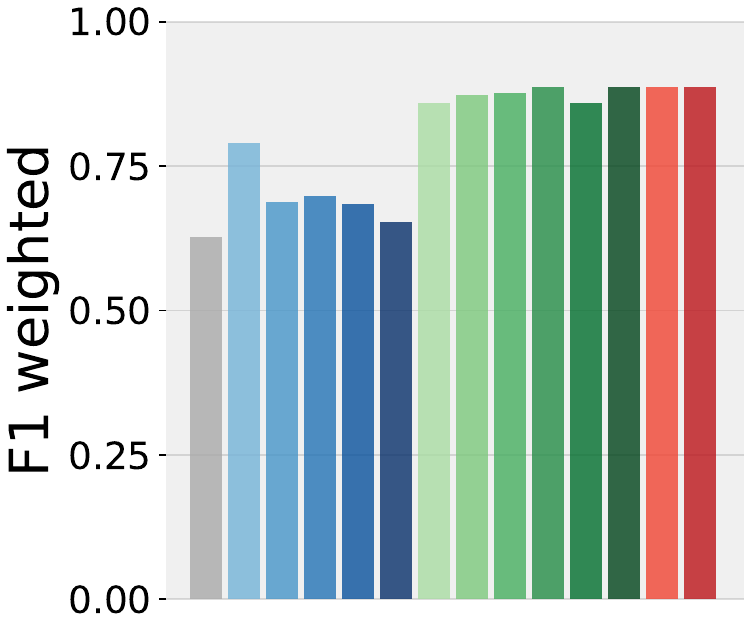}
        \caption{}
        \label{fig:sst2_short}
    \end{subfigure}
    \hfill
    \begin{subfigure}{0.235\textwidth} 
        \includegraphics[width=\textwidth]{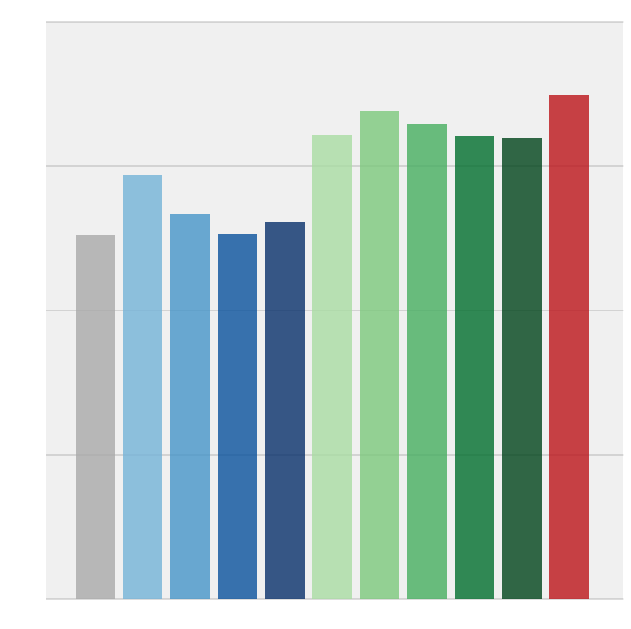}
        \caption{}
        \label{fig:sst2}
    \end{subfigure}
    \hfill
    \begin{subfigure}{0.235\textwidth}
        \includegraphics[width=\textwidth]{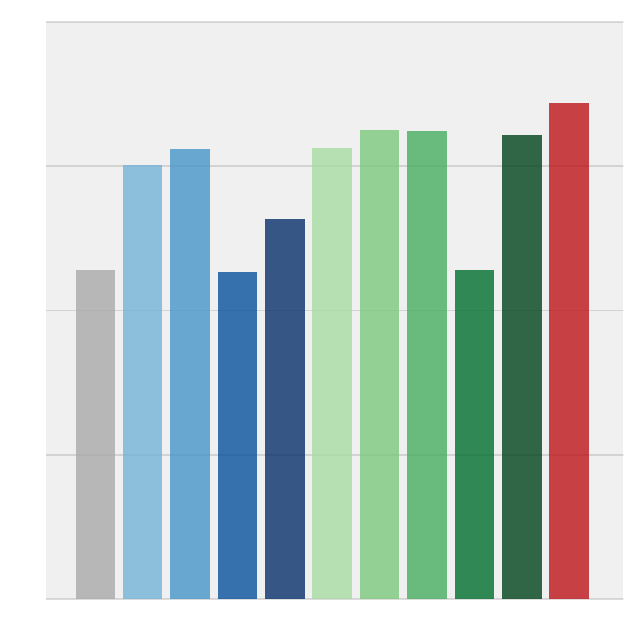}
        \caption{}
        \label{fig:imdb}
    \end{subfigure}
    \hfill
    \begin{subfigure}{0.235\textwidth}
        \includegraphics[width=\textwidth]{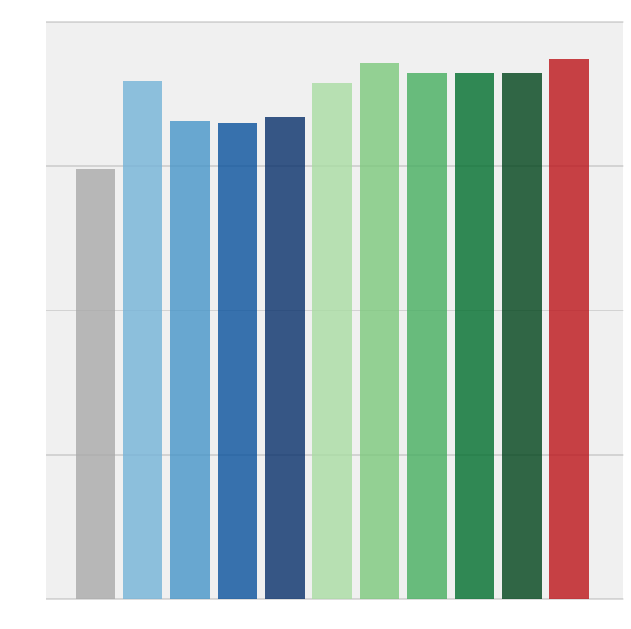}
        \caption{}
        \label{fig:ag_news}
    \end{subfigure}

    \caption{The F1 weighted scores across all methods for datasets (a) SST-2 short, (b) SST-2, (c) IMDb, and (d) Ag News. Methods are colour-coded such that blue corresponds to attention-based, green to gradient-based, and red to input-based methods.}
    \label{fig:nlp_barplot_f1}
\end{figure}

\begin{figure}[H]
    \centering
    \begin{subfigure}{\textwidth}\includegraphics[width=\textwidth]{Fig1a.pdf}
    \end{subfigure}
    \\
    \begin{subfigure}{0.285\textwidth} 
        \includegraphics[width=\textwidth]{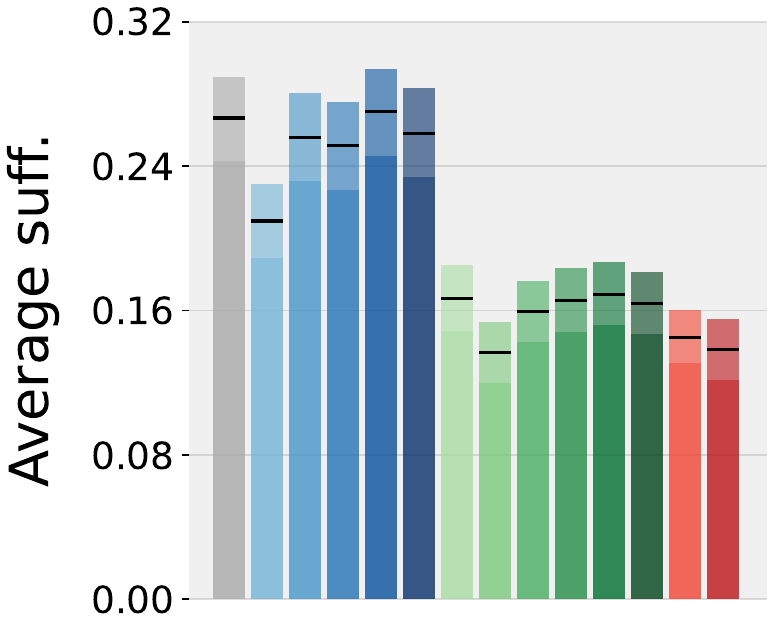}
        \caption{}
        \label{fig:sst2_short}
    \end{subfigure}
    \hfill
    \begin{subfigure}{0.233\textwidth} 
        \includegraphics[width=\textwidth]{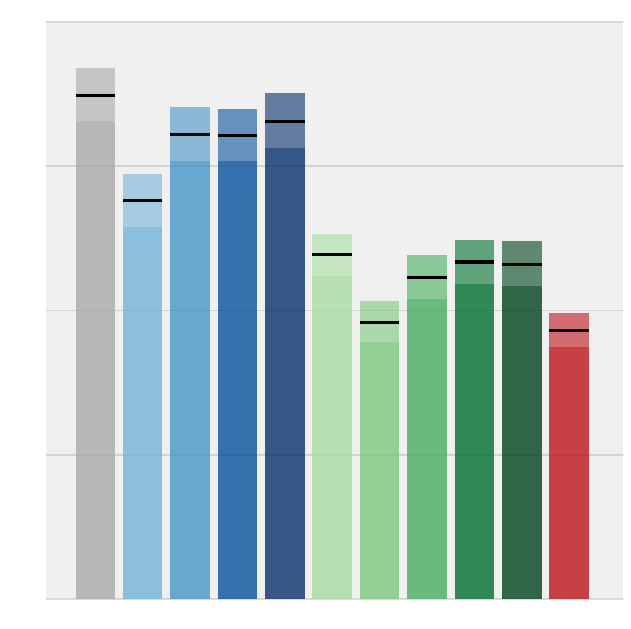}
        \caption{}
        \label{fig:sst2}
    \end{subfigure}
    \hfill
    \begin{subfigure}{0.233\textwidth}
        \includegraphics[width=\textwidth]{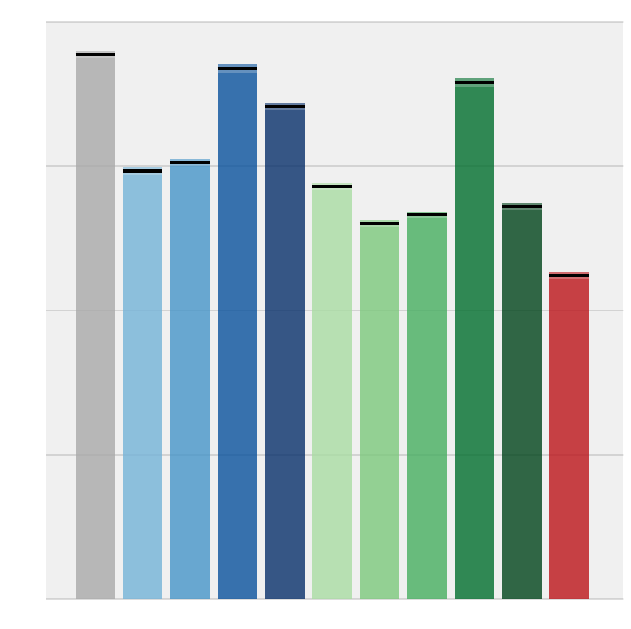}
        \caption{}
        \label{fig:imdb}
    \end{subfigure}
    \hfill
    \begin{subfigure}{0.233\textwidth}
        \includegraphics[width=\textwidth]{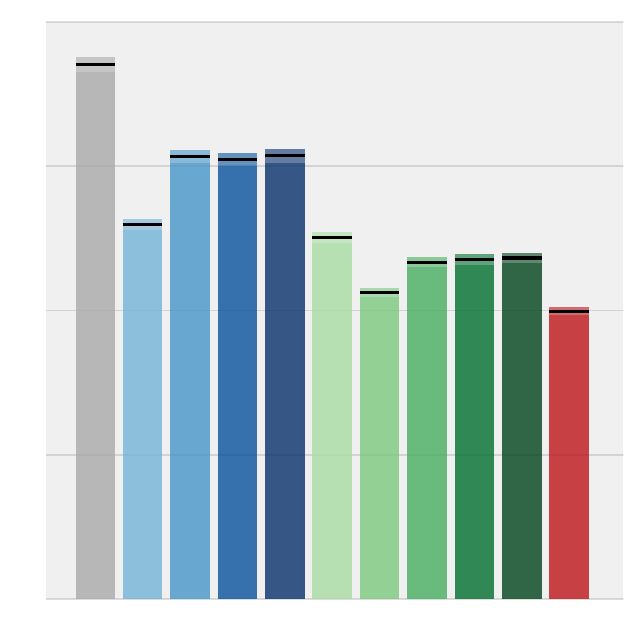}
        \caption{}
        \label{fig:ag_news}
    \end{subfigure}
    \caption{Average sufficiency (suff.) scores across all methods for datasets (a) SST-2 short, (b) SST-2, (c) IMDb, and (d) Ag News. The black horizontal lines mark the average values, and the lighter regions indicate the bootstrap confidence estimates. Methods are colour-coded such that blue corresponds to attention-based, green to gradient-based, and red to input-based methods.}
    \label{fig:nlp_barplot_suff}
\end{figure}

Results using metrics F1 weighted and sufficiency (see Sec.~\ref{sec:metrics}) for the NLP classification task across all methods and datasets are reported in Figs.~\ref{fig:nlp_barplot_f1} and \ref{fig:nlp_barplot_suff}, respectively. 

\FloatBarrier
\section{Qualitative comparison}
\label{app:qualitative_eval}

Table \ref{tab:nlp_qualitative_eval} provides a qualitative comparison of the methods Shapley-Grad-Att-CLS, Shapley-Grad-ATT-Mutual and SHAP, showing the top-20\% most important tokens as identified by each method for a subset of examples from the datasets SST-2 and Ag News. 

\begin{table*}[h]
\centering
\small
\begin{threeparttable}
\caption{Qualitative comparison of top-ranked tokens as identified by the methods Shapley-Grad-Att-CLS, Shapley-Grad-Att-Mutual, and SHAP, respectively.}
\label{tab:nlp_qualitative_eval}

\begin{tabularx}{\textwidth}{
>{\raggedright\arraybackslash}p{4.8cm}
>{\raggedright\arraybackslash}X
>{\raggedright\arraybackslash}X
>{\raggedright\arraybackslash}X}
\toprule
\textbf{Input (Dataset / Label)} & 
\textbf{Shapley-Grad-Att-CLS} & 
\textbf{Shapley-Grad-Att-Mutual} & 
\textbf{SHAP} \\
\midrule

\textit{SST / Positive} \newline
`easily my choice for one of the year's best films.'
& easily, best, films 
& best, easily, choice 
& best, films, one \\

\addlinespace[0.5em]

\textit{SST / Negative} \newline
rarely has so much money delivered so little entertainment. 
& little, so 
& little, entertainment 
& little, money \\

\addlinespace[0.5em]

\textit{SST / Negative} \newline
complete lack of originality, cleverness or even visible effort 
& lack, complete, of 
& lack, clever, or 
& lack, or, even \\

\addlinespace[0.5em]

\textit{AG News / World} \newline
Bush says he will work with allies President Bush told a Thursday news conference he would continue to lead the United States in promoting freedom and democracy in the Middle East.
& president, middle, east, thursday, bush, ., allies 
& middle, east, president, allies, bush, bush, told 
& bush, freedom, east, democracy, says, allies, he \\

\addlinespace[0.5em]

\textit{AG News / Sci/Tech} \newline
SafeGuard Offers Easy Hard-Drive Protection Upgraded version of this encryption app adds plenty of tools for networked users.
& encryption, safeguard, this, drive, upgraded 
& safeguard, encryption, app, version, offers 
& users, encryption, safeguard, hard, offers \\

\addlinespace[0.5em]

\textit{AG News / Business} \newline
Retail Sales Fall 0.3 Percent in August Retail sales slid in August as people steered away from buying cars and shoppers kept a close eye on their spending after splurging in July.
& retail, sales, retail, sales, .,  august, slid, fall 
& as, retail, slid, retail, sales, august, fall, sales
& sales, retail, retail, spending, sales, and, fall, shop \\

\bottomrule
\end{tabularx}
\end{threeparttable}
\end{table*}

\FloatBarrier
\newpage
\section{Variance of attention weights and concept sensitivity}

\begin{figure}[h]
    \centering
    \begin{subfigure}{\textwidth}
        \centering
        \includegraphics[width=0.5\textwidth]{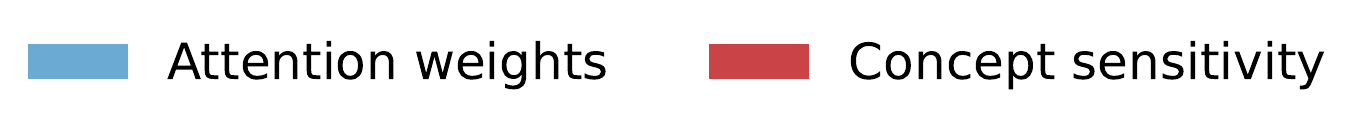}
    \end{subfigure}
    \\
    \begin{subfigure}{0.285\textwidth}
        \includegraphics[width=\textwidth]{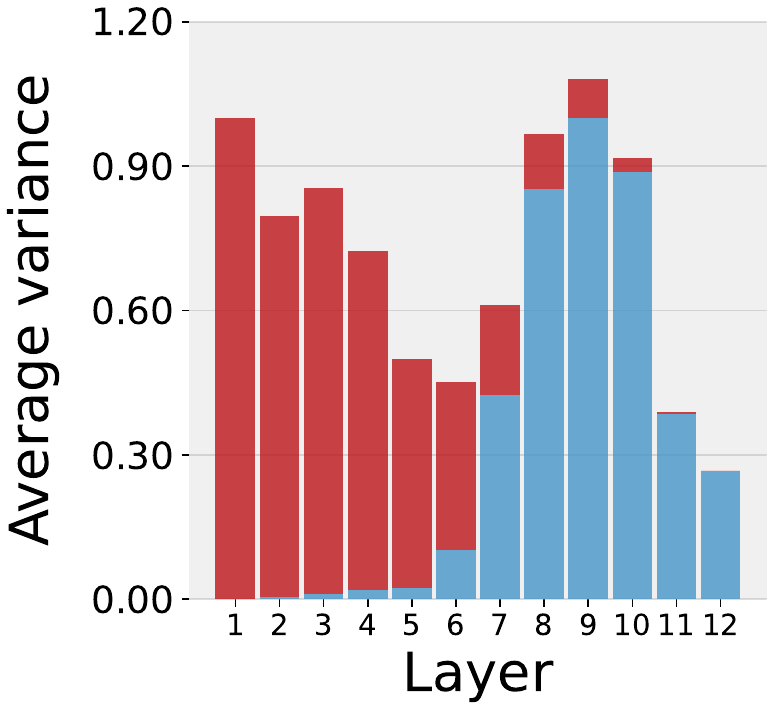}
        \caption{}
    \end{subfigure}
    \hfill
    \begin{subfigure}{0.233\textwidth}
        \includegraphics[width=\textwidth]{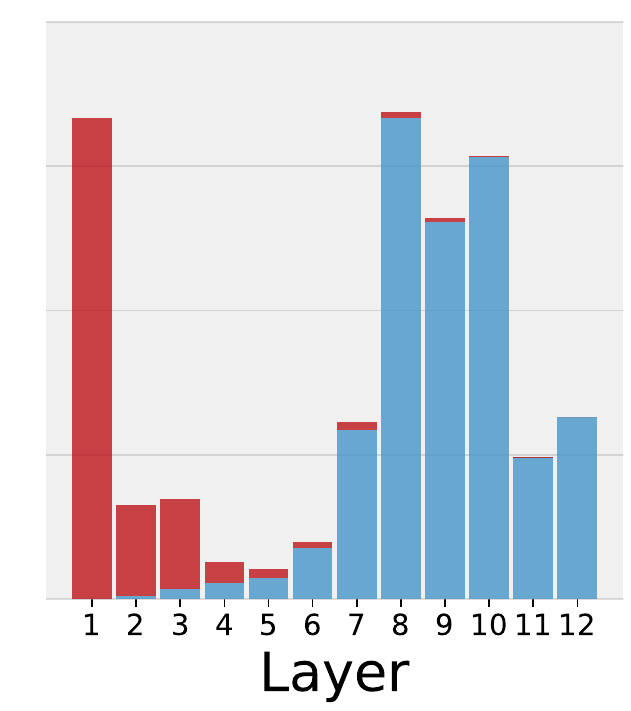}
        \caption{}
    \end{subfigure}
    \hfill
    \begin{subfigure}{0.233\textwidth}
        \includegraphics[width=\textwidth]{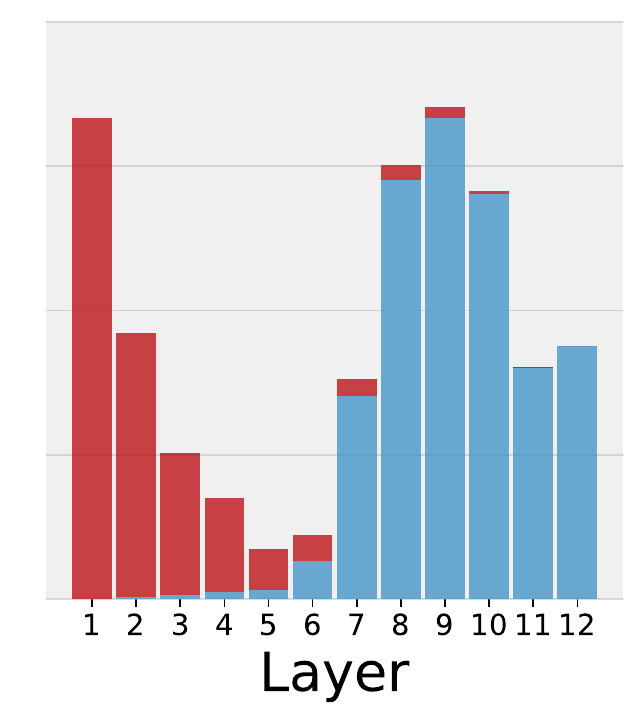}
        \caption{}
    \end{subfigure}
    \hfill
    \begin{subfigure}{0.233\textwidth}
        \includegraphics[width=\textwidth]{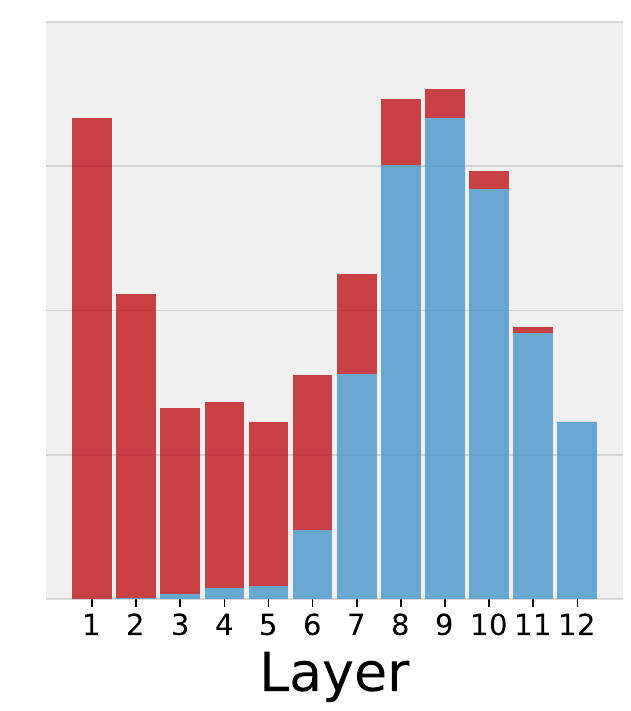}
        \caption{}
    \end{subfigure}
    
    \caption{Variance across tokens of (blue) the attention weights from the classification token to the image tokens (averaged over heads), and (red) the concept sensitivity. The latter is computed according to Eq.~\eqref{eq:token_directional_derivative}, without attention weighting. Results are averaged over multiple images and shown separately for the target classes (a) \textit{dalmatian}, (b) \textit{dog sled}, (c) \textit{fire engine}, and (d) \textit{zebra}. The self-attention weight of the classification token is excluded, and values are normalised to a common scale for comparability. }
    \label{fig:variance_all_classes}
\end{figure}

\begin{figure}[h]
    \centering
    
    \begin{subfigure}{0.285\textwidth}
        \centering
        \includegraphics[width=\textwidth]{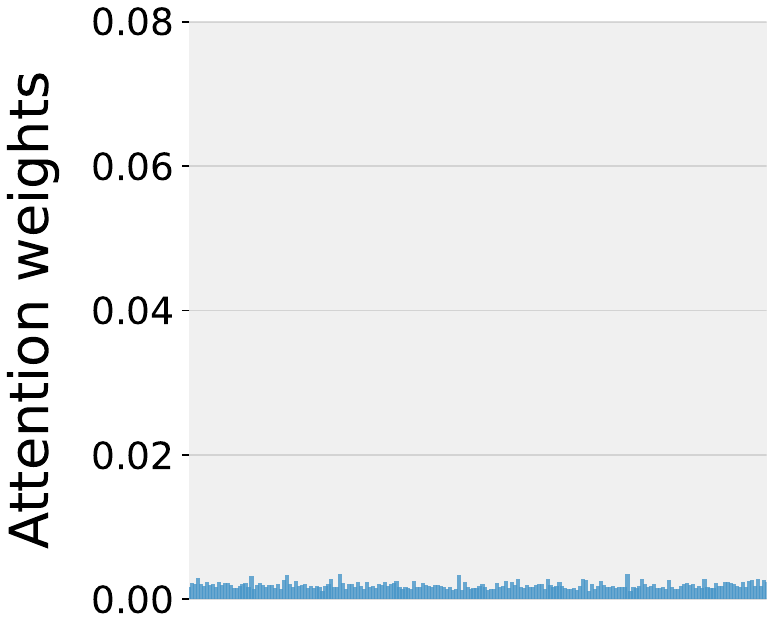}
        \caption{}
    \end{subfigure}
    \hfill
    \begin{subfigure}{0.233\textwidth}
        \centering
        \includegraphics[width=\textwidth]{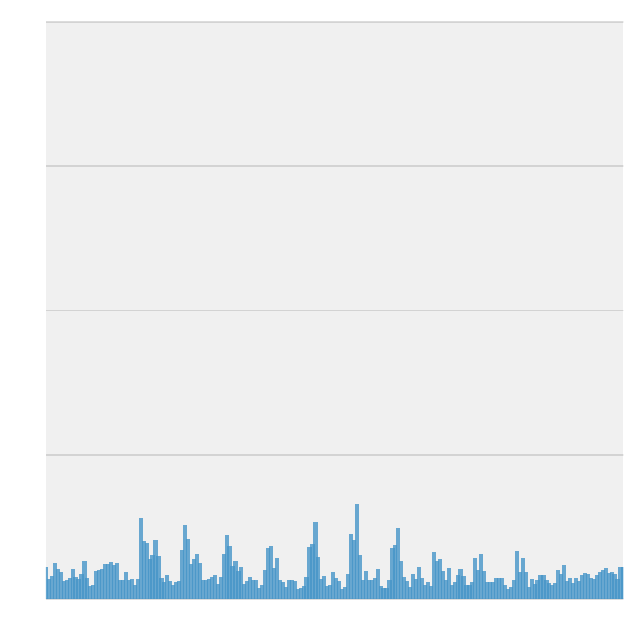}
        \caption{}
    \end{subfigure}
    \hfill
    \begin{subfigure}{0.233\textwidth}
        \centering
        \includegraphics[width=\textwidth]{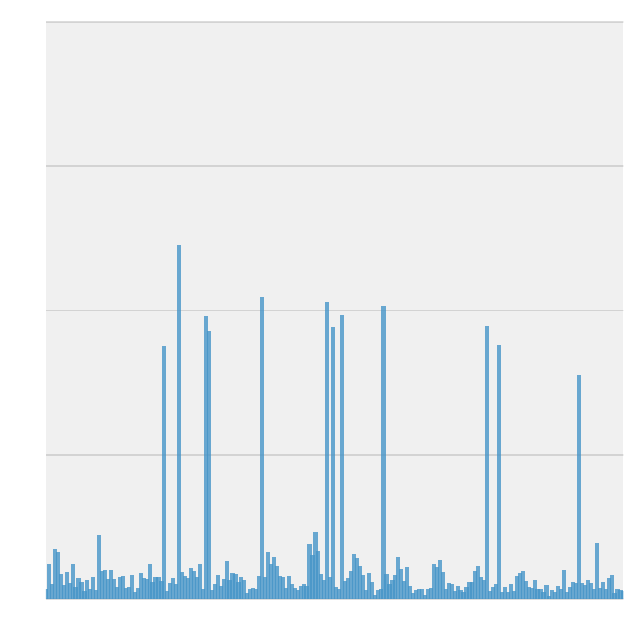}
        \caption{}
    \end{subfigure}
    \hfill
    \begin{subfigure}{0.233\textwidth}
        \centering
        \includegraphics[width=\textwidth]{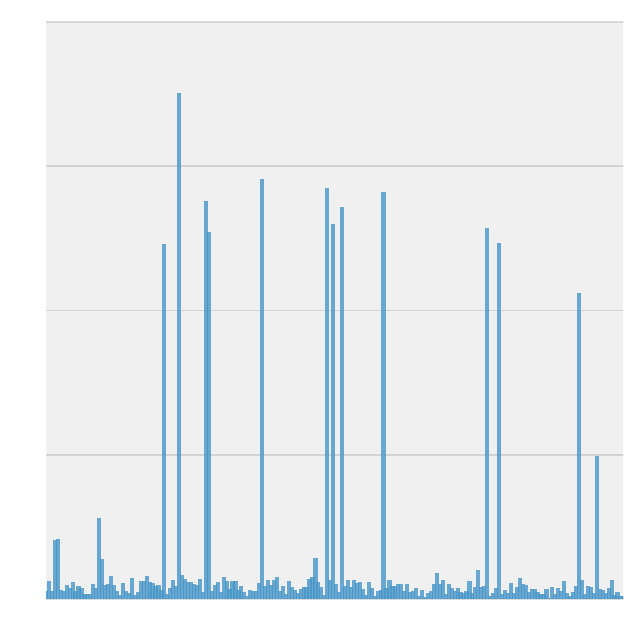}
        \caption{}
    \end{subfigure}
    
    \vspace{0.4cm}

    \begin{subfigure}{0.285\textwidth}
        \centering
        \includegraphics[width=\textwidth]{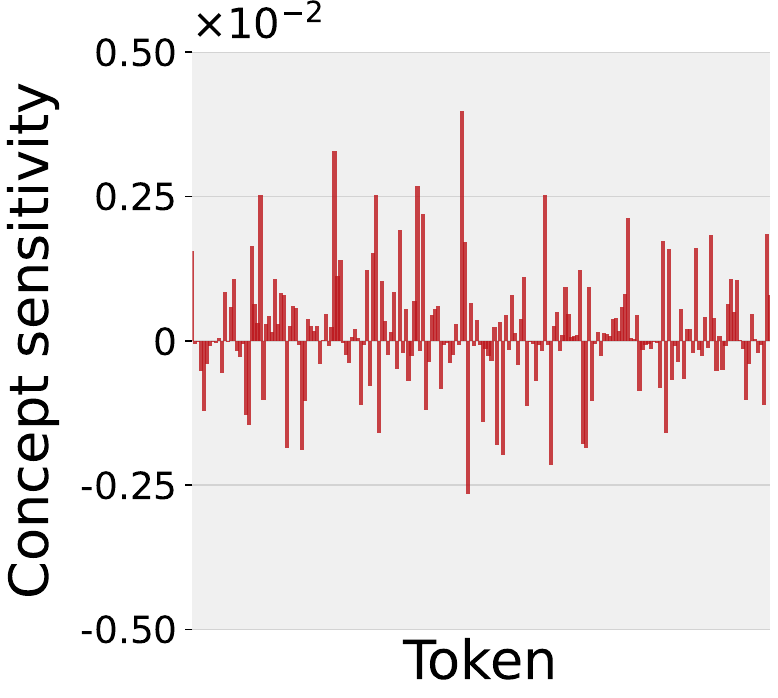}
        \caption{}
    \end{subfigure}
    \hfill
    \begin{subfigure}{0.233\textwidth}
        \centering
        \includegraphics[width=\textwidth]{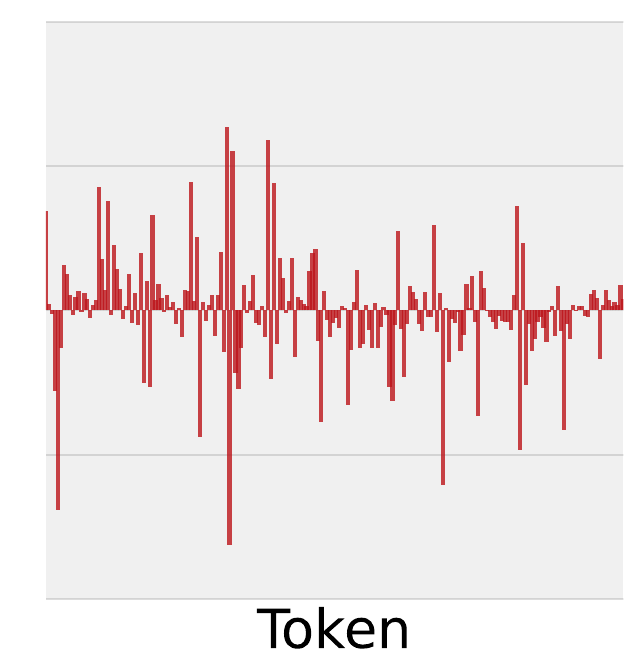}
        \caption{}
    \end{subfigure}
    \hfill
    \begin{subfigure}{0.233\textwidth}
        \centering
        \includegraphics[width=\textwidth]{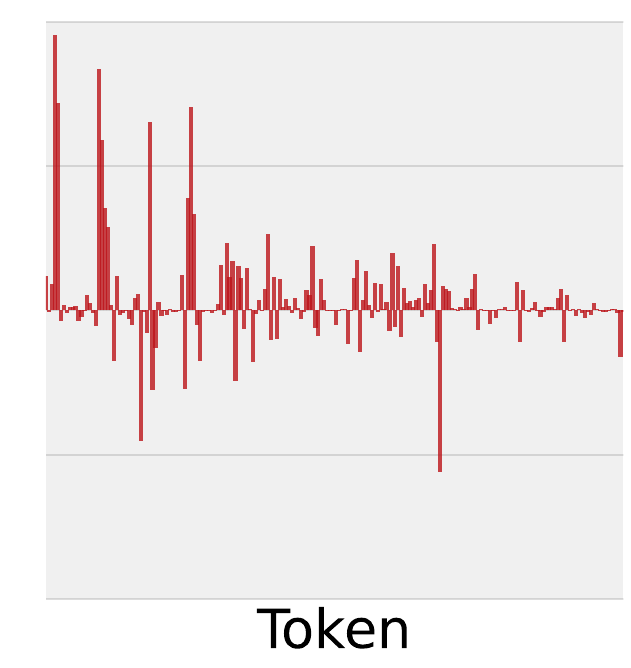}
        \caption{}
    \end{subfigure}
    \hfill
    \begin{subfigure}{0.233\textwidth}
        \centering
        \includegraphics[width=\textwidth]{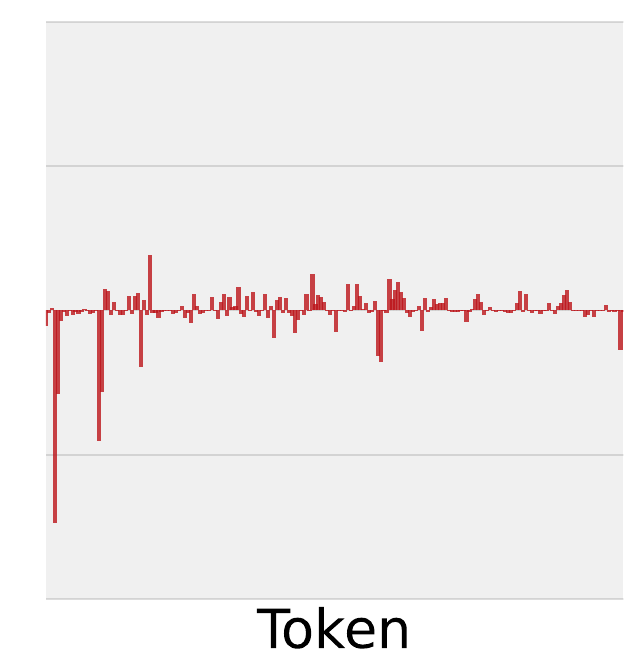}
        \caption{}
    \end{subfigure}
    \caption{The attention weights from the classification token to the image tokens (averaged over heads) and the concept sensitivity, both across tokens. The latter is computed according to Eq.~\eqref{eq:token_directional_derivative} without attention weighting. Results are shown for a randomly selected image from the \textit{dalmatian} class for layers (a, e) 1, (b, f) 3, (c, g) 7 and (d, h) 10. The self-attention weight of the classification token is excluded.}
    \label{fig:variance_dalmatian}
\end{figure}

Figure \ref{fig:variance_all_classes} shows the average variance across tokens of both the concept sensitivities and the attention weights from the classification token to the image tokens. The same quantities for a single image from the \textit{dalmatian} class are shown in Fig.~\ref{fig:variance_dalmatian}.

\FloatBarrier
\section{Comparison of T-TCAV and TCAV}

\begin{figure}[h]
    \centering
    \begin{subfigure}{\textwidth}
        \centering
        \includegraphics[width=0.25\textwidth]{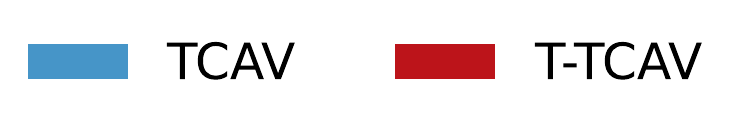}
    \end{subfigure}
    \\

    \begin{subfigure}{\textwidth}
        \includegraphics[width=\textwidth]{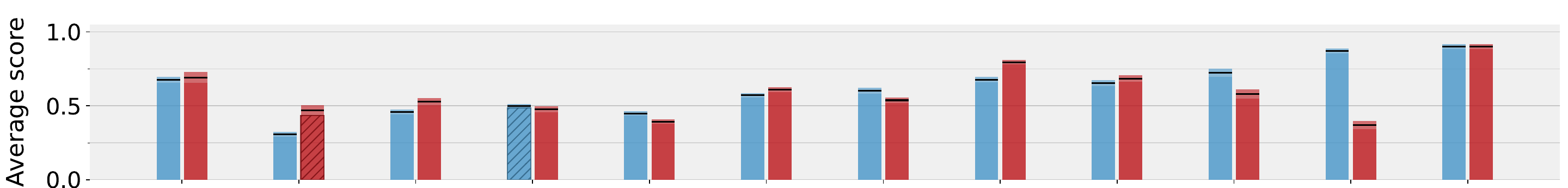}
        \caption{}
    \end{subfigure}
    \vfill

    \begin{subfigure}{\textwidth}
        \includegraphics[width=\textwidth]{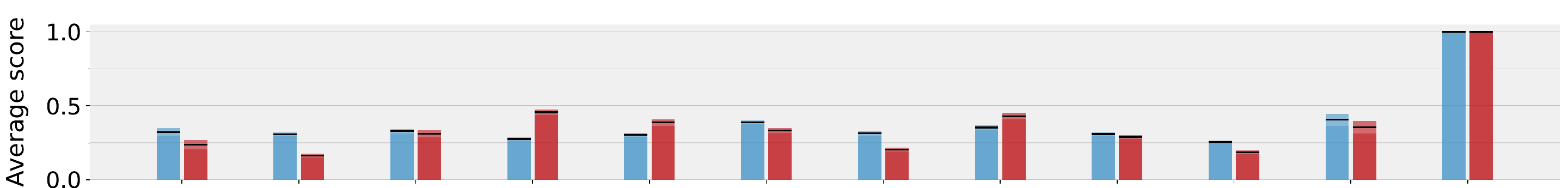}
        \caption{}
    \end{subfigure}
    \vfill  

    \begin{subfigure}{\textwidth}
        \includegraphics[width=\textwidth]{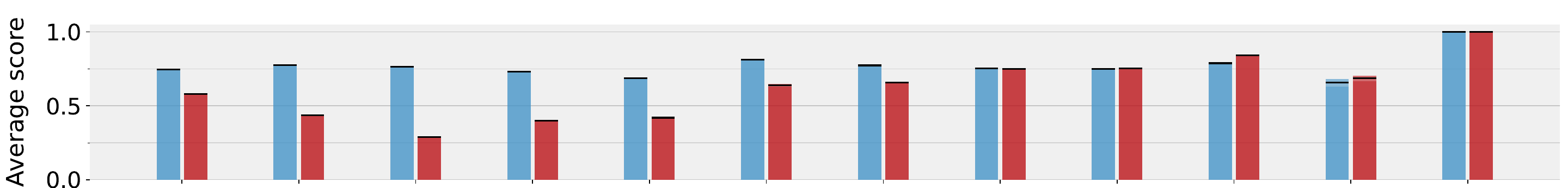}
        \caption{}
    \end{subfigure}
    \vfill

    \begin{subfigure}{\textwidth}
        \includegraphics[width=\textwidth]{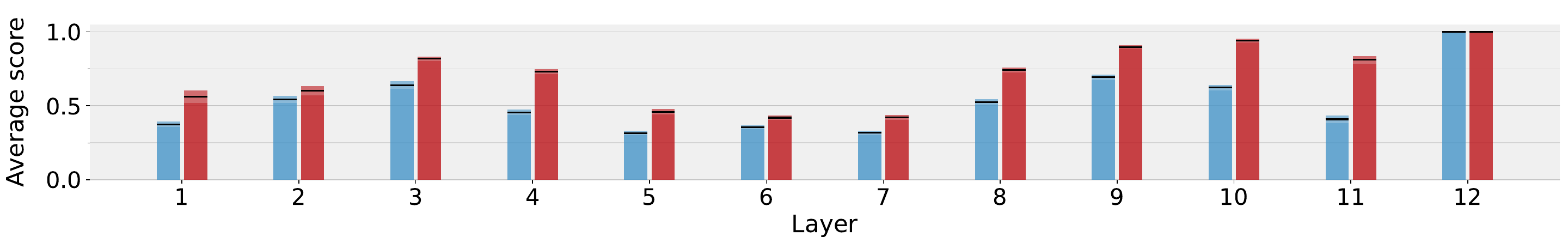}
        \caption{}
    \end{subfigure}

    \caption{The average TCAV and T-TCAV scores computed across 50 relative CAVs for the target classes (a) \textit{dalmatian}, (b) \textit{dog sled}, (c) \textit{fire engine}, and (d) \textit{zebra}, evaluated for concepts \textit{dotted}, \textit{siberian husky}, \textit{red}, and \textit{striped}, respectively. The lighter bar regions indicate confidence intervals, and bars with a hatched pattern mark scores that are not statistically significant according to a two-sided t-test, under the null hypothesis that the true score equals 0.5.}
    \label{fig:tcav_vs_ttcav}
\end{figure}

Figure \ref{fig:tcav_vs_ttcav} shows the proposed T-TCAV score compared to the TCAV method by \citet{pmlr-v80-kim18d}, both computed under identical experimental conditions. 

\end{document}